\pdfoutput=1
\documentclass{article}

    \PassOptionsToPackage{numbers, compress}{natbib}
\usepackage[final]{neurips_2022}

\usepackage[utf8]{inputenc} % allow utf-8 input
\usepackage[T1]{fontenc}    % use 8-bit T1 fonts
\usepackage{hyperref}       % hyperlinks
\usepackage{url}            % simple URL typesetting
\usepackage{booktabs}       % professional-quality tables
\usepackage{amsfonts}       % blackboard math symbols
\usepackage{nicefrac}       % compact symbols for 1/2, etc.
\usepackage{microtype}      % microtypography
\usepackage{xcolor}         % colors
\usepackage{graphicx}
\usepackage{amsthm,amsmath,amssymb}
\usepackage{algorithmic}
\usepackage{multirow}
\usepackage{color}
\usepackage[ruled, linesnumbered]{algorithm2e}

\title{A Win-win Deal: Towards Sparse and Robust Pre-trained Language Models}

\author{%
  Yuanxin Liu\textsuperscript{1,2,3}\thanks{Work was done when Yuanxin Liu was a graduate student of
IIE, CAS.}, Fandong Meng\textsuperscript{5}, Zheng Lin\textsuperscript{1,4}\thanks{Corresponding author: Zheng Lin.}, Jiangnan Li\textsuperscript{1,4}, Peng Fu\textsuperscript{1}, Yanan Cao\textsuperscript{1,4}, \\
  \textbf{Weiping Wang\textsuperscript{1}, Jie Zhou\textsuperscript{5}}\\
  \textsuperscript{1}Institute of Information Engineering, Chinese Academy of Sciences\\
  \textsuperscript{2}MOE Key Laboratory of Computational Linguistics, Peking University\\
  \textsuperscript{3}School of Computer Science, Peking University\\
  \textsuperscript{4}School of Cyber Security, University of Chinese Academy of Sciences\\
  \textsuperscript{5}Pattern Recognition Center, WeChat AI, Tencent Inc, China\thanks{Joint work with Pattern Recognition Center, WeChat AI, Tencent Inc, China.} \\
  {\tt liuyuanxin@stu.pku.edu.cn, \{fandongmeng,withtomzhou\}@tencent.com} \\
  {\tt \{linzheng,lijiangnan,fupeng,caoyanan,wangweiping\}@iie.ac.cn}\\
}

\begin{document}
\maketitle
\begin{abstract}
Despite the remarkable success of pre-trained language models (PLMs), they still face two challenges: First, large-scale PLMs are inefficient in terms of memory footprint and computation. Second, on the downstream tasks, PLMs tend to rely on the dataset bias and struggle to generalize to out-of-distribution (OOD) data. In response to the efficiency problem, recent studies show that dense PLMs can be replaced with sparse subnetworks without hurting the performance. Such subnetworks can be found in three scenarios: 1) the fine-tuned PLMs, 2) the raw PLMs and then fine-tuned in isolation, and even inside 3) PLMs without any parameter fine-tuning. However, these results are only obtained in the in-distribution (ID) setting. In this paper, we extend the study on PLMs subnetworks to the OOD setting, investigating whether sparsity and robustness to dataset bias can be achieved simultaneously. To this end, we conduct extensive experiments with the pre-trained BERT model on three natural language understanding (NLU) tasks. Our results demonstrate that \textbf{sparse and robust subnetworks (SRNets) can consistently be found in BERT}, across the aforementioned three scenarios, using different training and compression methods. Furthermore, we explore the upper bound of SRNets using the OOD information and show that \textbf{there exist sparse and almost unbiased BERT subnetworks}. Finally, we present 1) an analytical study that provides insights on how to promote the efficiency of SRNets searching process and 2) a solution to improve subnetworks' performance at high sparsity. The code is available at \url{https://github.com/llyx97/sparse-and-robust-PLM}.
\end{abstract}

\section{Introduction}
Pre-trained language models (PLMs) have enjoyed impressive success in natural language processing (NLP) tasks. However, they still face two major problems. On the one hand, the prohibitive model size of PLMs leads to poor efficiency in terms of memory footprint and computational cost \cite{ganesh-etal-2021-compressing,EnergyPolicyConsiderationNLP}. On the other hand, despite being pre-trained on large-scale corpus, PLMs still tend to rely on \textit{dataset bias} \cite{MNLI-hard,HANS,PAWS,FeverSym}, i.e., the spurious features of input examples that strongly correlate with the label, during downstream fine-tuning. These two problems pose great challenge to the real-world deployment of PLMs, and they have triggered two separate lines of works.

\begin{figure*}[t]
\centering
\includegraphics[width=0.8\textwidth]{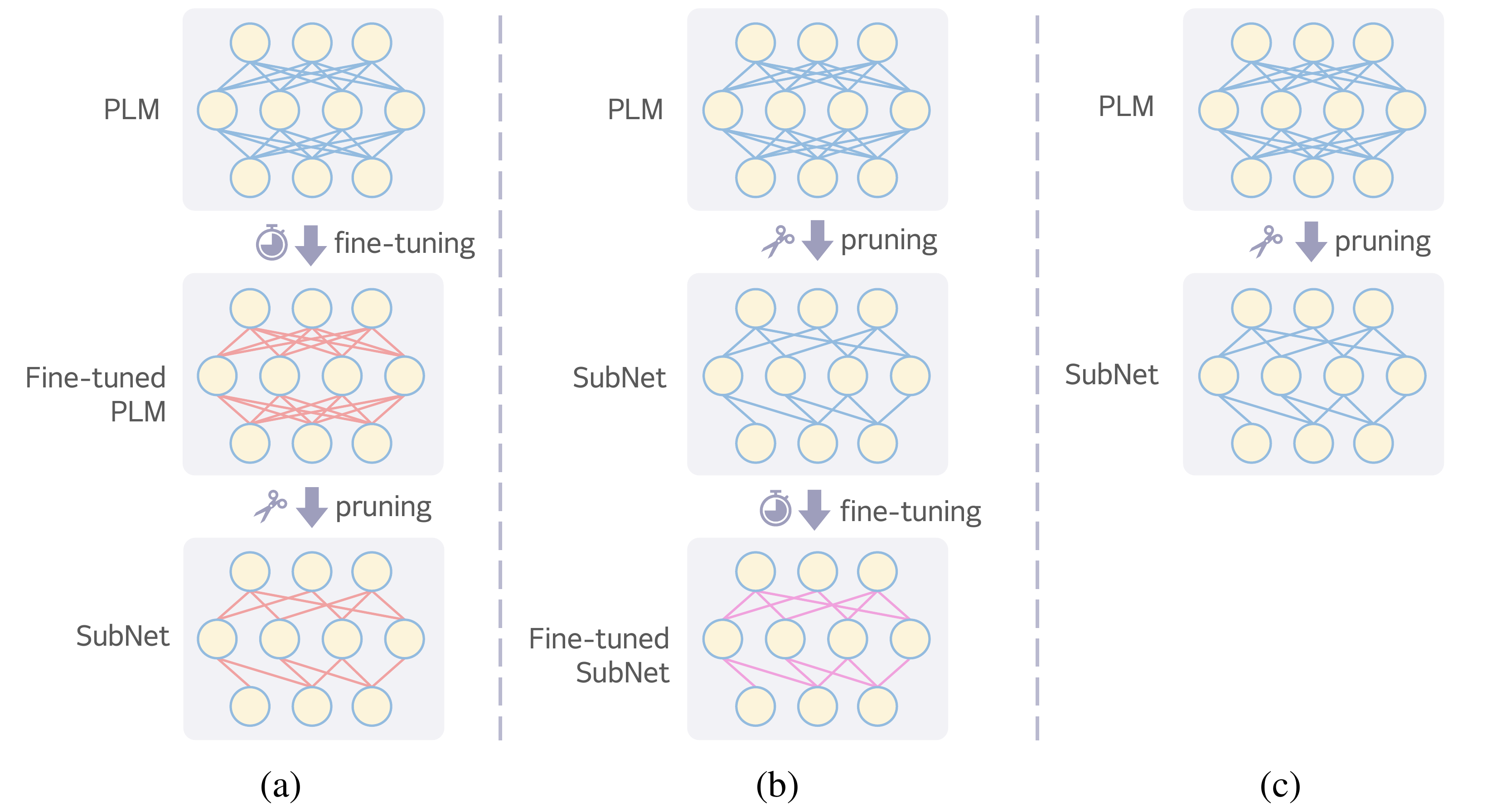}
\caption{Three kinds of PLM subnetworks obtained from different pruning and fine-tuning paradigms. (a) Pruning a fine-tuned PLM. (b) Pruning the PLM and then fine-tuning the subnetwork. (c) Pruning the PLM without fine-tuning model parameters. The obtained subnetworks are used for testing.}
\label{fig:sparsity}
\end{figure*}

In terms of the efficiency problem, some recent studies resort to sparse subnetworks as alternatives to the dense PLMs. \cite{TrainLargeThenCompress,AttnPrun,rosita} compress the fine-tuned PLMs in a post-hoc fashion. \cite{BERT-LT,AllTicketWin,TAMT,SuperTickets} extend the \textit{Lottery Ticket Hypothesis} (LTH) \cite{LTH} to search PLMs subnetworks that can be fine-tuned in isolation. Taking one step further, \cite{Masking} propose to learn task-specific subnetwork structures via mask training \cite{BinaryNet,Piggyback}, without fine-tuning any pre-trained parameter. Fig. \ref{fig:sparsity} illustrates these three paradigms. Encouragingly, the empirical evidences suggest that PLMs can indeed be replaced with sparse subnetworks without compromising the in-distribution (ID) performance.

To address the dataset bias problem, numerous debiasing methods have been proposed. A prevailing category of debiasing methods \cite{DontTakeEasyWay,MindTradeOff,EndToEndBias,UnlearnDatasetBias,FeverSym,EndToEndSelfDebias,UnknownBias} adjust the importance of training examples, in terms of training loss, according to their bias degree, so as to reduce the impact of biased examples (examples that can be correctly classified based on the spurious features). As a result, the model is forced to rely less on the dataset bias during training and generalizes better to OOD situations.

%%%%%%%%%%%%%%%%%%%%%%%%%%%%%%%%%%%%%%%%%%%%%%
%%%%%%%%%%%%%%%%%%%%%%%%%%%%%%%%%%%%%%%%%%%%%%

Although progress has been made in both directions, most existing work tackle the two problems independently. To facilitate real-world application of PLMs, the problems of robustness and efficiency should be addressed simultaneously. Motivated by this, we extend the study on PLM subnetwork to the OOD scenario, investigating \textbf{whether there exist PLM subnetworks that are both sparse and robust against dataset bias?} To answer this question, we conduct large-scale experiments with the pre-trained BERT model \cite{BERT} on three natural language understanding (NLU) tasks that are widely-studied in the question of dataset bias. We consider a variety of setups including the three pruning and fine-tuning paradigms, standard and debiasing training objectives, different model pruning methods, and different variants of PLMs from the BERT family. Our results show that \textbf{BERT does contain sparse and robust subnetworks (SRNets)} within certain sparsity constraint (e.g., less than 70\%), giving affirmative answer to the above question. Compared with a standard fine-tuned BERT, SRNets exhibit comparable ID performance and remarkable OOD improvement. When it comes to BERT model fine-tuned with debiasing method, SRNets can preserve the full model's ID and OOD performance with much fewer parameters. On this basis, we further explore the upper bound of SRNets by making use of the OOD information, which reveals that \textbf{there exist sparse and almost unbiased subnetworks, even in a standard fine-tuned BERT that is biased}.

Regardless of the intriguing properties of SRNets, we find that the subnetwork searching process still have room for improvement, based on some observations from the above experiments. First, we study the timing to start searching SRNets during full BERT fine-tuning, and find that the entire training and searching cost can be reduced from this perspective. Second, we refine the mask training method 
with gradual sparsity increase, which is quite effective in identifying SRNets at high sparsity.

Our main contributions are summarized as follows: 
\begin{itemize}
\item We extend the study on PLMs subnetworks to the OOD scenario. To our knowledge, this paper presents the first systematic study on sparsity and dataset bias robustness for PLMs.
\item We conduct extensive experiments to demonstrate the existence of sparse and robust BERT subnetworks, across different pruning and fine-tuning setups. By using the OOD information, we further reveal that there exist sparse and almost unbiased BERT subenetworks.
\item We present analytical studies and solutions that can help further refine the SRNets searching process in terms of efficiency and the performance of subnetworks at high sparsity.
\end{itemize}

\section{Related Work}
\subsection{BERT Compression}
Studies on BERT compression can be divided into two classes. The first one focuses on the design of model compression techniques, which include pruning \cite{CompressingBERT,AttnPrun,StateofSparsity}, knowledge distillation \cite{DistillBERT,PKD,TinyBERT,MUD}, parameter sharing \cite{ALBERT}, quantization \cite{Q8BERT,TernaryBERT}, and combining multiple techniques \cite{EdgeBERT,LadaBERT,rosita}. The second one, which is based on the lottery ticket hypothesis \cite{LTH}, investigates the compressibility of BERT on different phases of the pre-training and fine-tuning paradigm. It has been shown that BERT can be pruned to a sparse subnetwork after \cite{StateofSparsity} and before fine-tuning \cite{BERT-LT,AllTicketWin,SuperTickets,TAMT,CompressingBERT}, without hurting the accuracy. Moreover, \cite{Masking} show that directly learning subnetwork structures on the pre-trained weights can match fine-tuning the full BERT. In this paper, we follow the second branch of works, and extend the evaluation of BERT subnetworks to the OOD scenario.

\subsection{Dataset Bias in NLP Tasks}
To facilitate the development of NLP systems that truly learn the intended task solution, instead of relying on dataset bias, many efforts have been made recently. On the one hand, challenging OOD test sets are constructed \cite{MNLI-hard,HANS,PAWS,FeverSym,VQA-CP} by eliminating the spurious correlations in the training sets, in order to establish more strict evaluation. On the other hand, numerous debiasing methods \cite{DontTakeEasyWay,MindTradeOff,EndToEndBias,UnlearnDatasetBias,FeverSym,EndToEndSelfDebias,UnknownBias} are proposed to discourage the model from learning dataset bias during training. However, few attention has been paid to the influence of pruning on the OOD generalization ability of PLMs. This work presents a systematic study on this question.

\subsection{Model Compression and Robustness}
Some pioneer attempts have also been made to obtain models that are both compact and robust to adversarial attacks \cite{ADMM1,ADMM2,HYDRA,RobustScratchTickets,LoyaltyRobustnessOfBERT} and spurious correlations \cite{MRM,WhatDoCompressLMForget}. Specially, \cite{LoyaltyRobustnessOfBERT,WhatDoCompressLMForget} study the compression and robustness question on PLM. Different from \cite{LoyaltyRobustnessOfBERT}, which is based on adversarial robustness, we focus on the spurious correlations, which is more common than the worst-case adversarial attack. Compared with \cite{WhatDoCompressLMForget}, which focus on post-hoc pruning of the standard fine-tuned BERT, we thoroughly investigate different fine-tuning methods (standard and debiasing) and subnetworks obtained from the three pruning and fine-tuning paradigms. A more detailed discussion of the relation and difference between our work and previous studies on model compression and robustness is provided in Appendix \ref{appendix-d}.

\section{Preliminaries}
\subsection{BERT Architecture and Subnetworks}
\label{sec:bert_struc_and_subnet}
BERT is composed of an embedding layer, a stack of Transformer layers \cite{Transformer} and a task-specific classifier. Each Transformer layer has a multi-head self-attention (MHAtt) module and a feed-forward network (FFN). MHAtt has four kinds of weight matrices, i.e., the query, key and value matrices $\mathbf{W}_{Q,K,V} \in \mathbb{R}^{d_{\text{model}} \times d_{\text{model}}}$, and the output matrix $\mathbf{W}_{AO} \in \mathbb{R}^{d_{\text{model}} \times d_{\text{model}}}$. FFN consits of two linear layers $\mathbf{W}_{\text{in}} \in \mathbb{R}^{d_{\text{model}} \times d_{\text{FFN}}}$, $\mathbf{W}_{\text{out}} \in \mathbb{R}^{d_{\text{FFN}} \times d_{\text{model}}}$, where $d_{\text{FFN}}$ is the hidden dimension of FFN.

To obtain the subnetwork of a model $f(\boldsymbol{\boldsymbol{\theta}})$ parameterized by $\boldsymbol{\theta}$, we apply a binary pruning mask $\mathbf{m} \in\{0,1\}^{|\boldsymbol{\theta}|}$ to its weight matrices, which produces $f(\mathbf{m} \odot \boldsymbol{\theta})$, where $\odot$ is the Hadamard product. For BERT, we focus on the $L$ Transformer layers and the classifier. The parameters to be pruned are $\boldsymbol{\theta}_{pr} = \{\mathbf{W}_{\text{cls}}\} \cup \left\{\mathbf{W}_{Q}^{l}, \mathbf{W}_{K}^{l}, \mathbf{W}_{V}^{l}, \mathbf{W}_{A O}^{l}, \mathbf{W}_{\text{in}}^{l}, \mathbf{W}_{\text{out}}^{l}\right\}_{l=1}^{L}$, where $\mathbf{W}_{\text{cls}}$ is the classifier weights.

\subsection{Pruning Methods}
\subsubsection{Magnitude-based Pruning}
\label{sec:magnitude_pruning}
Magnitude-based pruning \cite{HanNips,LTH} zeros-out parameters with low absolute values. It is usually realized in an iterative manner, namely, iterative magnitude pruning (IMP). IMP alternates between pruning and training and gradually increases the sparsity of subnetworks. Specifically, a typical IMP algorithm consists of four steps: (i) Training the full model to convergence. (ii) Pruning a fraction of parameters with the smallest magnitude. (iii) Re-training the pruned subnetwork. (iv) Repeat (ii)-(iii) until reaching the target sparsity. To obtain subnetworks from the pre-trained BERT, i.e., (b) and (c) in Fig. \ref{fig:sparsity}, the subnetwork parameters are rewound to the pre-trained values after (iii), and (i) can be abandoned. More details about our IMP implementations can be found in Appendix \ref{appendix-a1.1}.

\subsubsection{Mask Training}
\label{sec:mask_train}
Mask training treats the pruning mask $\mathbf{m}$ as trainable parameters. Following \cite{Piggyback,Masking,HowFine,TAMT}, we achieve this through binarization in forward pass and gradient estimation in backward pass.

Each weight matrix $\mathbf{W} \in \mathbb{R}^{d_1 \times d_2}$, which is frozen during mask training, is associated with a bianry mask $\mathbf{m} \in \{0,1\}^{d_1 \times d_2}$, and a real-valued mask $\hat{\mathbf{m}} \in \mathbb{R}^{d_1 \times d_2}$. In the forward pass, $\mathbf{W}$ is replaced with $\mathbf{m} \odot \mathbf{W}$, where $\mathbf{m}$ is derived from $\hat{\mathbf{m}}$ through binarization:
\begin{equation}
\label{eq:binarization}
\mathbf{m}_{i, j}= \begin{cases}1 & \text { if } \hat{\mathbf{m}}_{i, j} \geq \phi \\ 0 & \text { otherwise }\end{cases}
\end{equation}
where $\phi$ is the threshold. In the backward pass, since the binarization operation is not differentiable, we use the \textit{straight-through estimator} \cite{GradEstimator} to compute the gradients for $\hat{\mathbf{m}}$ using the gradients of $\mathbf{m}$, i.e., $\frac{\partial \mathcal{L}}{\partial \mathbf{m}}$, where $\mathcal{L}$ is the loss. Then, $\hat{\mathbf{m}}$ is updated as $\hat{\mathbf{m}} \leftarrow \hat{\mathbf{m}}-\eta \frac{\partial \mathcal{L}}{\partial \mathbf{m}}$, where $\eta$ is the learning rate.

Following \cite{HowFine,TAMT}, we initialize the real-valued masks according to the magnitude of the original weights. The complete mask training algorithm is summarized in Appendix \ref{appendix-a1.2}.

\subsection{Debiasing Methods}
\label{sec:debiasing}
As described in the Introduction, the debiasing methods measure the bias degree of training examples. This is achieved by training a \textit{bias model}. The inputs to the bias model are hand-crafted spurious features based on our prior knowledge of the dataset bias (Section \ref{sec:train_details} describes the details). In this way, the bias model mainly relies on the spurious features to make predictions, which can then serve as a measurement of the bias degree. Specifically, given the bias model prediction $\mathbf{p}_b = (\mathbf{p}^{1}_b, \cdots, \mathbf{p}^{K}_b)$ over the $K$ classes, the bias degree $\beta = \mathbf{p}^{c}_b$, i.e., the the probability of the ground-truth class $c$. 

Then, $\beta$ can be used to adjust the training loss in several ways, including \textit{product-of-experts} (PoE) \cite{DontTakeEasyWay,UnlearnDatasetBias,EndToEndBias}, \textit{example reweighting} \cite{FeverSym,EndToEndSelfDebias} and \textit{confidence regularization} \cite{MindTradeOff}. Here we describe the standard cross-entropy and PoE, and the other two methods are introduced in Appendix \ref{appendix-a2}.

\textbf{Standard Cross-Entropy} computes the cross-entropy between the predicted distribution $\mathbf{p}_m$ and the ground-truth one-hot distribution $\mathbf{y}$ as $\mathcal{L}_{\text{std}}=-\mathbf{y} \cdot \log \mathbf{p}_m$.

\textbf{Product-of-Experts} combines the predictions of main model and bias model, i.e., $\mathbf{p}_b$ and $\mathbf{p}_m$, and then computes the training loss as $\mathcal{L}_{\text{poe}}=-\mathbf{y} \cdot \log \operatorname{softmax}\left(\log \mathbf{p}_m+\log \mathbf{p}_b\right)$.

\subsection{Notations}
Here we define some notations, which will be used in the following sections.
\begin{itemize}
\item $\mathcal{A}^{t}_{\mathcal{L}}(f(\boldsymbol{\theta}))$: Training $f(\boldsymbol{\theta})$ with loss $\mathcal{L}$ for $t$ steps, where $t$ can be omitted for simplicity.

\item $\mathcal{P}^{p}_{\mathcal{L}}(f(\boldsymbol{\theta}))$: Pruning $f(\boldsymbol{\theta})$ using pruning method $p$ and training loss $\mathcal{L}$.

\item $\mathcal{M}(f(\mathbf{m}\boldsymbol{\theta}))$: Extracting the pruning mask of $f(\mathbf{m}\boldsymbol{\theta})$, i.e., $\mathcal{M}(f(\mathbf{m}\boldsymbol{\theta})) = \mathbf{m}$.

\item $\mathcal{L} \in \{\mathcal{L}_{\text{std}}, \mathcal{L}_{\text{poe}}, \mathcal{L}_{\text{reweight}}, \mathcal{L}_{\text{confreg}}\}$ and $p \in \{\text{imp}, \text{imp-rw}, \text{mask}\}$, where ``imp'' and ``imp-rw''denote the standard IMP and IMP with weight rewinding, as described in Section \ref{sec:magnitude_pruning}. ``mask'' stands for mask training.

\item $\mathcal{E}_d(f(\boldsymbol{\theta}))$: Evaluating $f(\boldsymbol{\theta})$ on the test data with distribution $d \in \{\text{ID}, \text{OOD}\}$.
\end{itemize}

\section{Sparse and Robust BERT Subnetworks}
\label{sec:srnets}
\subsection{Experimental Setups}
\label{sec:experi_setups}

\subsubsection{Datasets and Evaluation}
\paragraph{Natural Language Inference} We use MNLI \cite{MNLI} as the ID dataset for NLI. MNLI is comprised of premise-hypothesis pairs, whose relationship may be \textit{entailment}, \textit{contradiction}, or \textit{neutral}. In MNLI the word overlap between premise and hypothesis is strongly correlated with the \textit{entailment} class. To solve this problem, the OOD HANS dataset \cite{HANS} is built so that such correlation does not hold.

\paragraph{Paraphrase Identification} The ID dataset for paraphrase identification is QQP \footnote{\url{https://www.kaggle.com/c/quora-question-pairs}}, which contains question pairs that are labelled as either \textit{duplicate} or \textit{non-duplicate}. In QQP, high lexical overlap is also strongly associated with the \textit{duplicate} class. The OOD datasets PAWS-qqp and PAWS-wiki \cite{PAWS} are built from sentences in Quora and Wikipedia respectively. In PAWS sentence pairs with high word overlap have a balanced distribution over \textit{duplicate} and \textit{non-duplicate}.

\paragraph{Fact Verification} FEVER \footnote{See the licence information at \url{https://fever.ai/download/fever/license.html}} \cite{FEVER} is adopted as the ID dataset of fact verification, where the task is to assess whether a given evidence \textit{supports} or \textit{refutes} the claim, or whether there is \textit{not-enough-info} to reach a conclusion. The OOD dataset Fever-Symmetric (v1 and v2) \cite{FeverSym} is proposed to evaluate the influence of the claim-only bias (the label can be predicted correctly without the evidence).

For NLI and fact verification, we use Accuracy as the evaluation metric. For paraphrase identification, we evaluate using the F1 score. More details of datasets and evaluation are shown in Appendix \ref{appendix-b1}.

\subsubsection{PLM Backbone}
\label{sec:setup_model}
We mainly experiment with the BERT-base-uncased model \cite{BERT}. It has roughly 110M parameters in total, and 84M parameters in the Transformer layers. As described in Section \ref{sec:bert_struc_and_subnet}, we derive the subnetworks from the Transformer layers and report sparsity levels relative to the 84M parameters. To generalize our conclusions to other PLMs, we also consider two variants of the BERT family, namely RoBERTa-base and BERT-large, the results of which can be found in Appendix \ref{appendix-c5}.

\subsubsection{Training Details}
\label{sec:train_details}
Following \cite{DontTakeEasyWay}, we use a simple linear classifier as the bias model. For HANS and PAWS, the spurious features are based on the the word overlapping information between the two input text sequences. For Fever-Symmetric, the spurious features are max-pooled word embeddings of the claim sentence. More details about the bias model and the spurious features are presented in Appendix \ref{appendix-b3.1}.

Mask training and IMP basically use the same hyper-parameters (adopting from \cite{UnknownBias}) as full BERT. An exception is longer training, because we find that good subnetworks at high sparsity levels require more training to be found. Unless otherwise specified, we select the best checkpoints based on the ID dev performance, without using OOD information. All the reported results are averaged over 4 runs. We defer training details about each dataset, and each training and pruning setup, to Appendix \ref{appendix-b3}.

\subsection{Subnetworks from Fine-tuned BERT}
\label{sec:after_ft}
\subsubsection{Problem Formulation and Experimental Setups}
Given the fine-tuned full BERT $f(\boldsymbol{\theta}_{ft}) = \mathcal{A}_{\mathcal{L}_1}(f(\boldsymbol{\theta}_{pt}))$, where $\boldsymbol{\theta}_{pt}$ and $\boldsymbol{\theta}_{ft}$ are the pre-trained and fine-tuned parameters respectively, the goal is to find a subnetwork $f(\mathbf{m} \odot \boldsymbol{\theta}^{'}_{ft}) = \mathcal{P}^{p}_{\mathcal{L}_2}(f(\boldsymbol{\theta}_{ft}))$ that satisfies a target sparsity level $s$ and maximize the ID and OOD performance.
\begin{equation}
\label{eq:after_ft}
\max _{\mathbf{m},\boldsymbol{\theta}^{'}_{ft}}\left(\mathcal{E}_{\text{ID}}\left(f \left( \mathbf{m} \odot \boldsymbol{\theta}^{'}_{ft}\right)\right) + \mathcal{E}_{\text{OOD}}\left(f \left( \mathbf{m} \odot \boldsymbol{\theta}^{'}_{ft}\right)\right) \right) ,
\text { s.t. } \frac{\|\mathbf{m}\|_{0}}{|\boldsymbol{\theta}_{pr}|} = (1-s)
\end{equation}
where $\|\|_{0}$ is the $L_0$ norm and $|\boldsymbol{\theta}_{pr}|$ is the total number of parameters to be pruned. In practice, the above optimization problem is achieved via $\mathcal{P}^{p}_{\mathcal{L}_2}()$, which minimizes the loss $\mathcal{L}_2$ on the ID training set. When the pruning method is IMP, the subnetwork parameters will be further fine-tuned and $\boldsymbol{\theta}^{'}_{ft} \neq \boldsymbol{\theta}_{ft}$. For mask training, only the subnetwork structure is updated and $\boldsymbol{\theta}^{'}_{ft} = \boldsymbol{\theta}_{ft}$.

We consider two kinds of fine-tuned full BERT, which utilize the standard CE loss and PoE loss respectively (i.e., $\mathcal{L}_1 \in \{\mathcal{L}_{\text{std}}, \mathcal{L}_{\text{poe}}\}$). IMP and mask training are used as the pruning methods (i.e., $p \in \{\text{imp}, \text{mask}\}$). For the standard fine-tuned BERT, both $\mathcal{L}_{\text{std}}$ and $\mathcal{L}_{\text{poe}}$ are examined in the pruning process. For the PoE fine-tuned BERT, we only use $\mathcal{L}_{\text{poe}}$ during pruning. Note that in this work, we mainly experiment with $\mathcal{L}_{\text{std}}$ and $\mathcal{L}_{\text{poe}}$. $\mathcal{L}_{\text{reweight}}$ and $\mathcal{L}_{\text{confreg}}$ are also examined for subnetworks from fine-tuned BERT, the results of which can be found in Appendix \ref{appendix-c1}.

%%%%%%%%%%%%%%%%%%%%%%%%%%%%%%%%%%%%%%%%%%%%%%
\begin{figure}
    \parbox{0.25\linewidth}{
        \centering
        \includegraphics[width=1.0\linewidth]{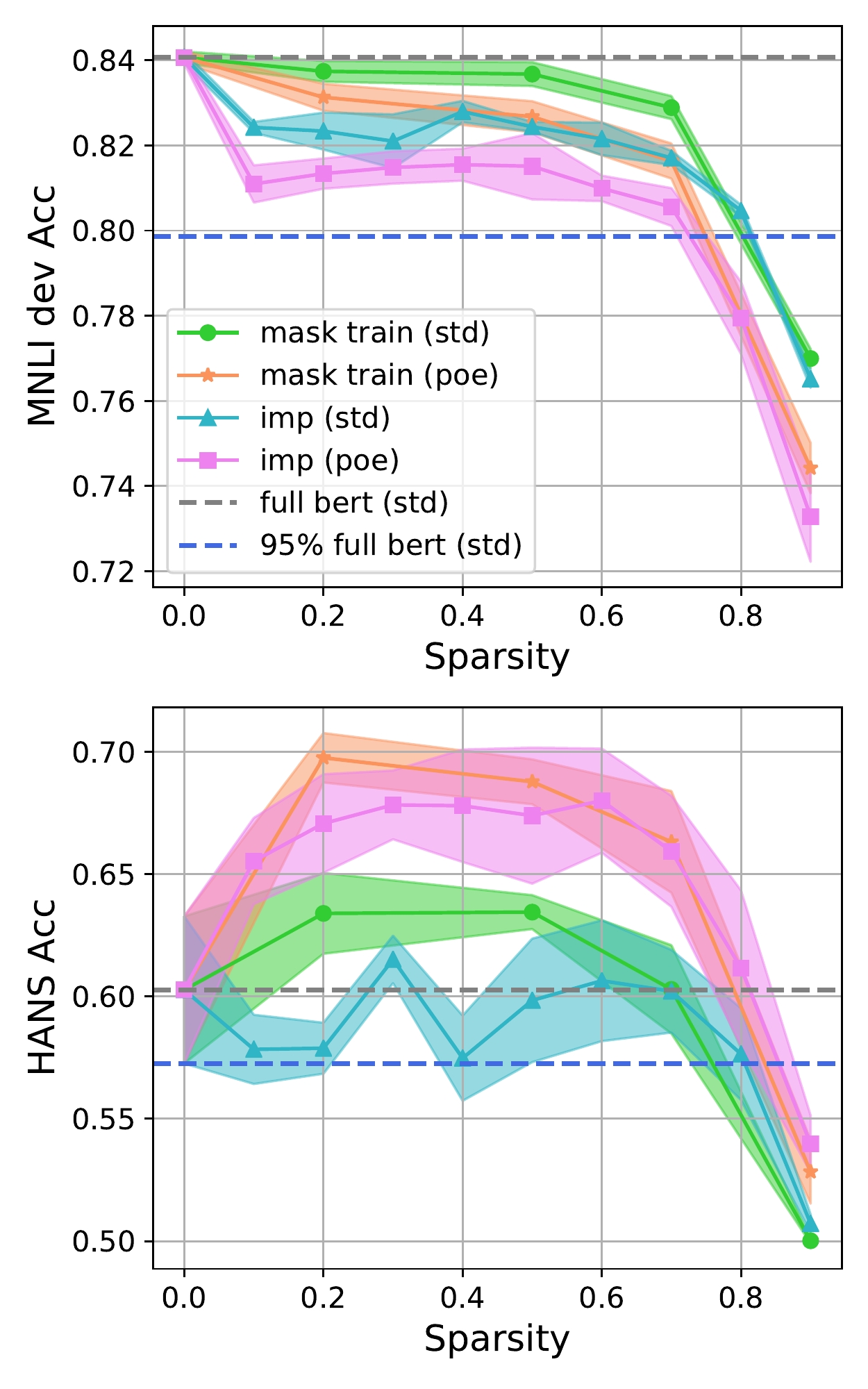}
    }
    \hfill
    \parbox{0.73\linewidth}{
        \centering
        \includegraphics[width=1.0\linewidth]{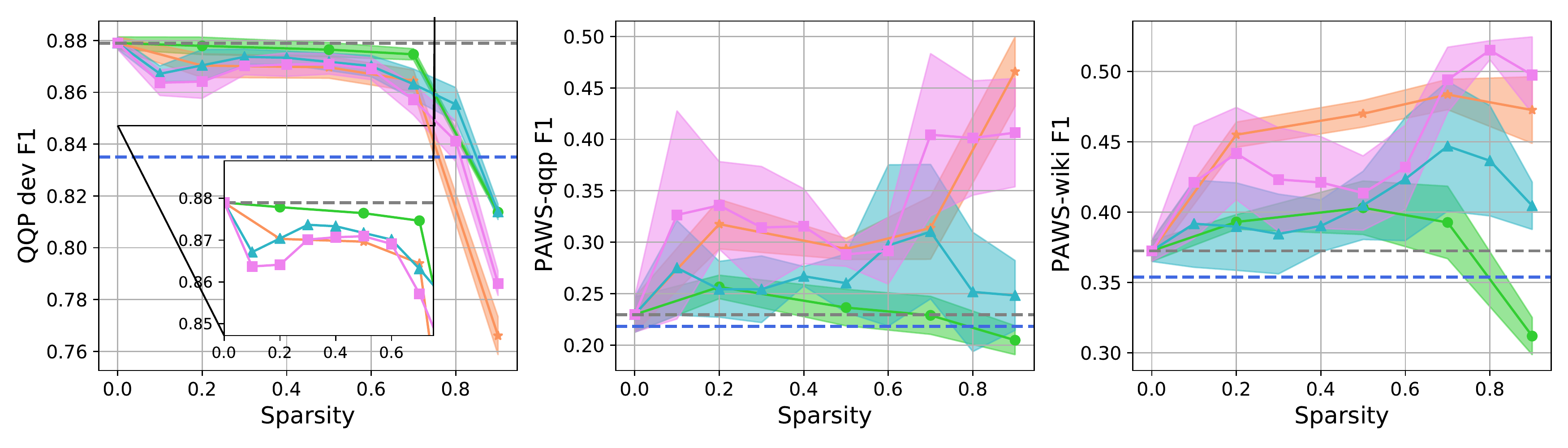}
        \includegraphics[width=1.0\linewidth]{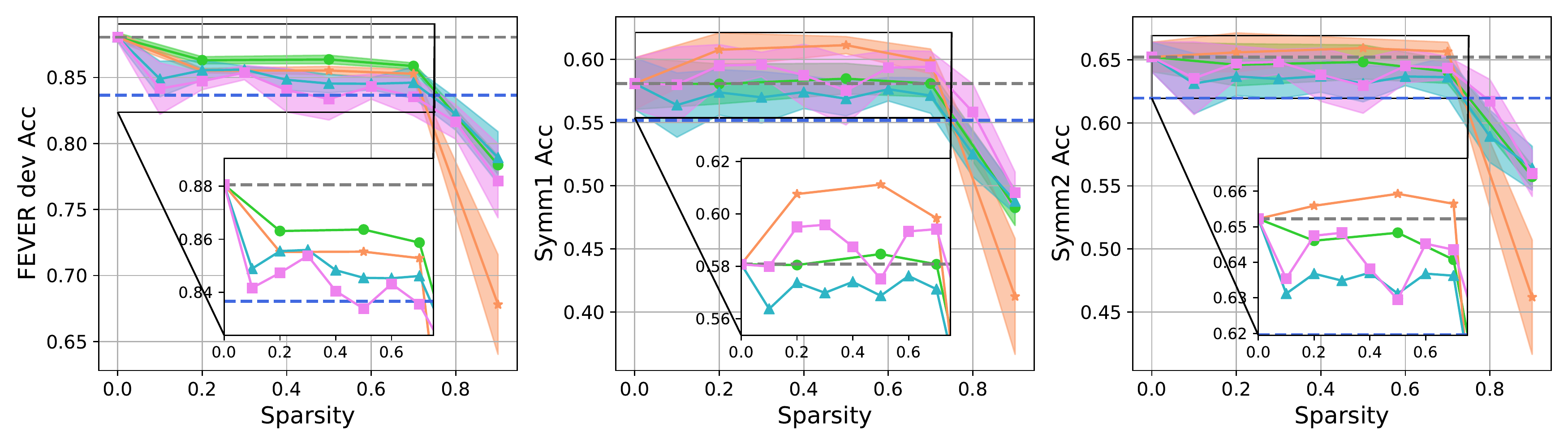}
    }
    \caption{Results of subnetworks pruned from the CE fine-tuned BERT. ``std'' means standard, and the shadowed areas denote standard deviations, which also apply to the other figures of this paper.}
    \label{fig:after_std_ft}
\end{figure}
%%%%%%%%%%%%%%%%%%%%%%%%%%%%%%%%%%%%%%%%%%%%%%
%%%%%%%%%%%%%%%%%%%%%%%%%%%%%%%%%%%%%%%%%%%%%%
\begin{figure}
    \parbox{0.25\linewidth}{
        \centering
        \includegraphics[width=1.0\linewidth]{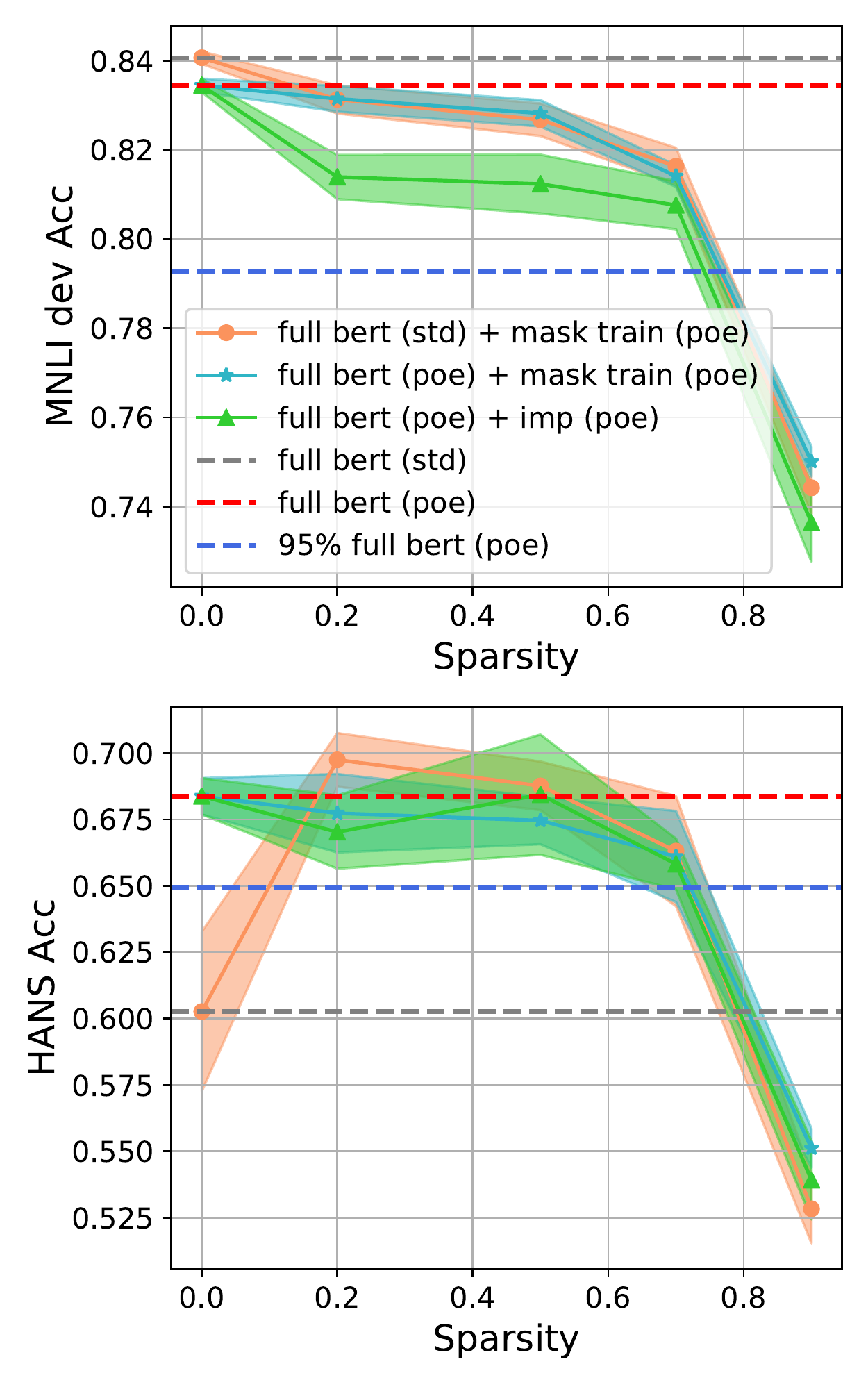}
    }
    \hfill
    \parbox{0.73\linewidth}{
        \centering
        \includegraphics[width=1.0\linewidth]{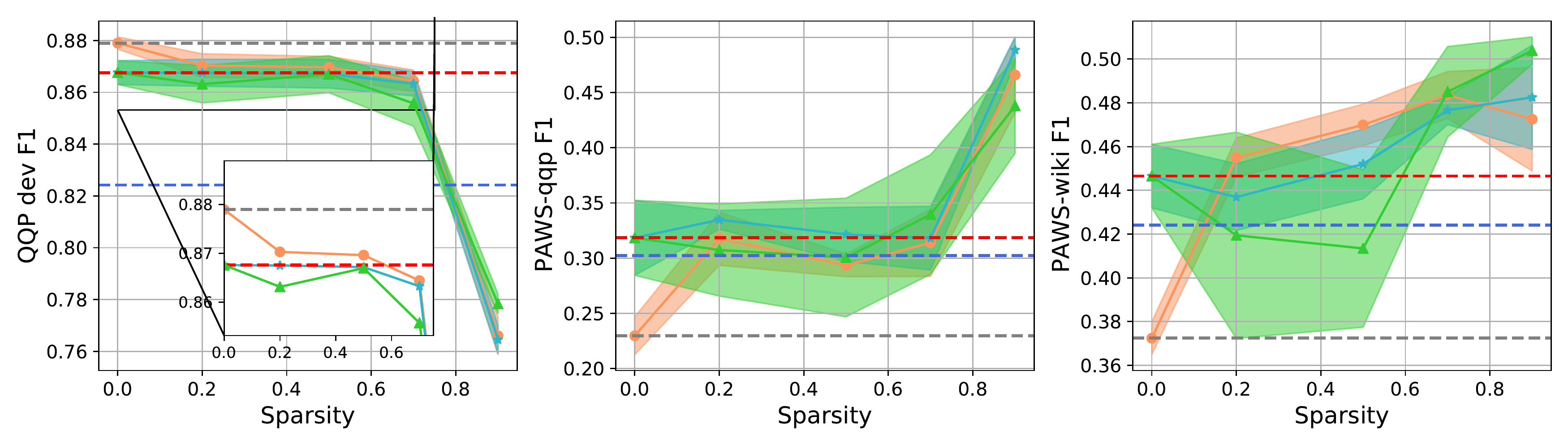}
        \includegraphics[width=1.0\linewidth]{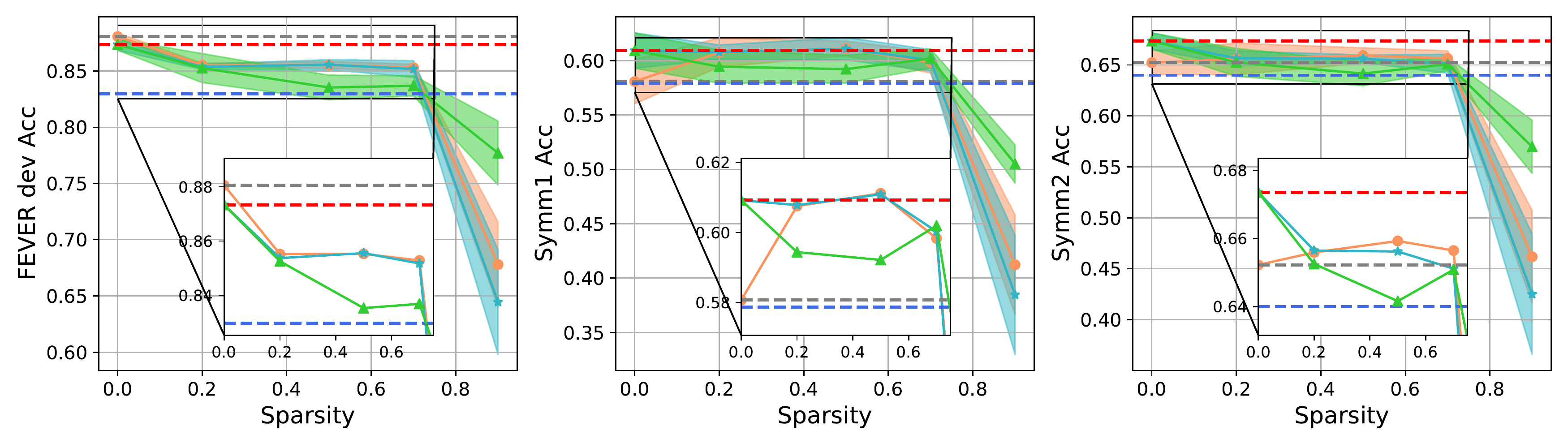}
    }
    \caption{Results of subnetworks pruned from the PoE fine-tuned BERT. Results of the ``mask train (poe)'' subnetworks from Fig. \ref{fig:after_std_ft} (the orange line) are also reported for reference.}
    \label{fig:after_poe_ft}
\end{figure}
%%%%%%%%%%%%%%%%%%%%%%%%%%%%%%%%%%%%%%%%%%%%%%

\subsubsection{Results}
\label{sec:result_after_ft}
\paragraph{Subnetworks from Standard Fine-tuned BERT} The results are shown in Fig. \ref{fig:after_std_ft} (In this paper, we present most results in figures for clear comparisons. Actual values of the results can be found in the code link.). We discuss them from three perspectives. For the full BERT, we can see that standard CE fine-tuning, which achieves good results on the ID dev sets, performs significantly worse on the OOD test sets. This demonstrates that the ID performance of BERT depends, to a large extent, on memorizing the dataset bias.

In terms of the subnetworks, we can derive the following observations: (1) Using any of the four pruning methods, we can compress a large proportion of the BERT parameters (up to $70\%$ sparsity) and still preserve $95\%$ of the full model's ID performance. (2)
With standard pruning, i.e., ``mask train (std)'' or ``imp (std)'', we can observe small but perceivable improvement over the full BERT on the HANS and PAWS datasets. This suggests that pruning may remove some parameters related to the bias features. (3) The OOD performance of ``mask train (poe)'' and ``imp (poe)'' subnetworks is even better, and the ID performance degrades slightly but is still above $95\%$ of the full BERT. This shows that introducing the debiasing objective in the pruning process is beneficial. Specially, as mask training does not change the model parameters, the results of ``mask train (poe)'' implicates that the biased ``full bert (std)'' contains sparse and robust subnetworks (SRNets) that already encode a less biased solution to the task. (4) SRNets can be identified across a wide range of sparsity levels (from $20\% \sim 70\%$). However at higher sparsity of $90\%$, the performance of the subnetworks is not desirable. (5) We also find that there is an abnormal increase of the PAWS F1 score at $70\% \sim 90\%$ sparsity for some pruning methods, when the corresponding ID performance drops sharply. This is because the class distribution of PAWS is imbalanced (see Appendix \ref{appendix-b1}), and thus even a naive random-guessing model can outperform the biased full model on PAWS. Therefore, the OOD improvement should only be acceptable when there is no large ID performance decline.

Comparing IMP and mask training, the latter performs better in general, except for ``mask train (poe)'' at $90\%$ sparsity on QQP and FEVER. This suggests that directly optimizing the subnetwork structure is a better choice than using the magnitude heuristic as the pruning metric.

\paragraph{Subnetworks from PoE Fine-tuned BERT} Fig. \ref{fig:after_poe_ft} presents the results. We can find that: (1) For the full BERT, the OOD performance is obviously promoted with the PoE debiasing method, while the ID performance is sacrificed slightly. (2) Unlike the subnetworks from the standard fine-tuned BERT, the subnetworks of PoE fine-tuned BERT (the green and blue lines) cannot outperform the full model. However, these subnetworks maintain comparable performance at up to $70\%$ sparsity, on both the ID and OOD settings, making them desirable alternatives to the full model in resource-constraint scenarios. Moreover, this phenomenon suggests that there is a great redundancy of BERT parameters, even when OOD generalization is taken into account. (3) With PoE-based pruning, subnetworks from the standard fine-tuned BERT (the orange line) is comparable with subnetworks from the PoE fine-tuned BERT (the blue line). This means we do not have to fine-tune a debiased BERT before searching for the SRNets. (4) IMP, again, slightly underperforms mask training at moderate sparsity levels, while it is better at $90\%$ sparsity on the fact verification task.

%%%%%%%%%%%%%%%%%%%%%%%%%%%%%%%%%%%%%%%%%%%%%%
\begin{figure}
    \parbox{0.25\linewidth}{
        \centering
        \includegraphics[width=1.0\linewidth]{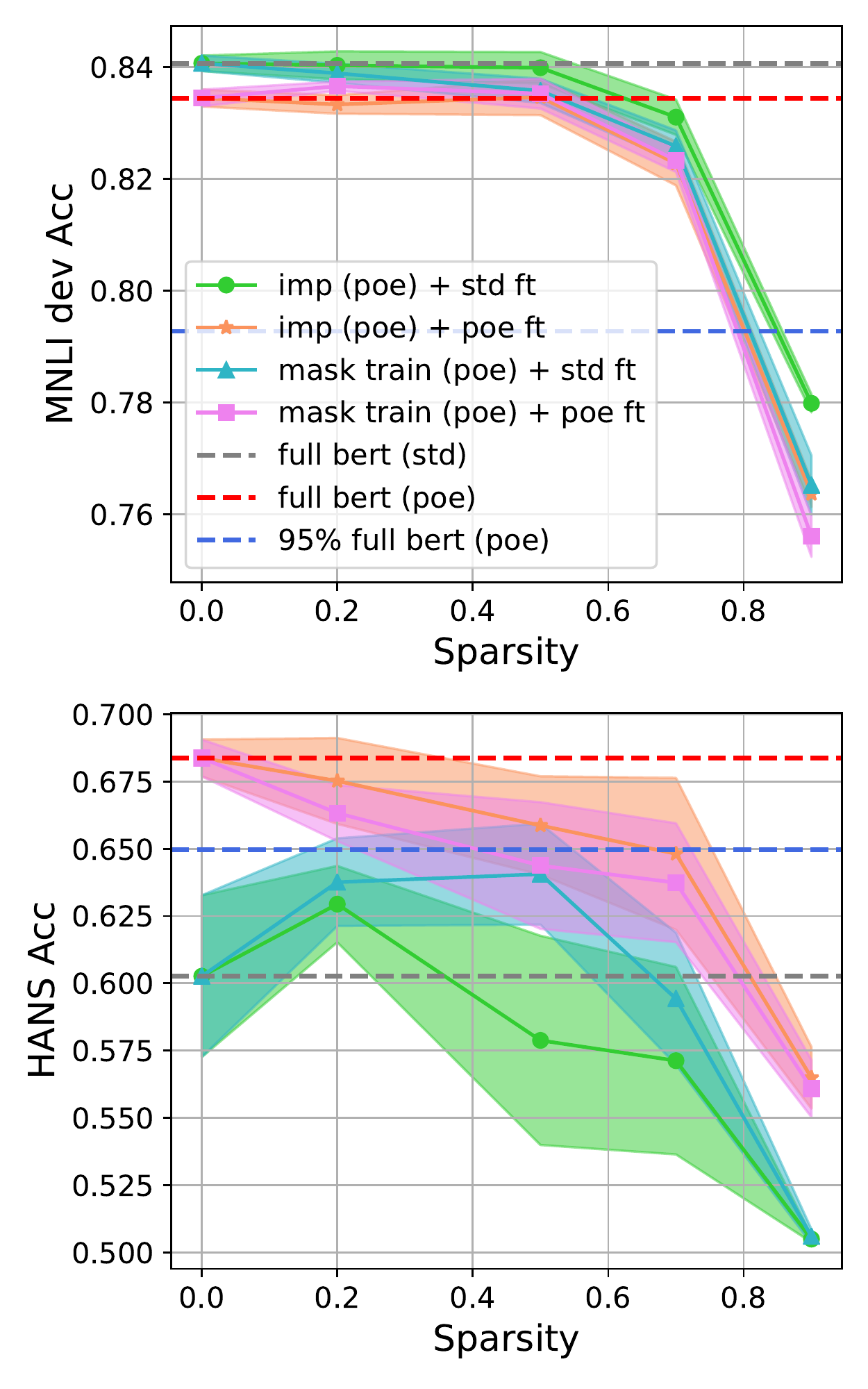}
    }
    \hfill
    \parbox{0.73\linewidth}{
        \centering
        \includegraphics[width=1.0\linewidth]{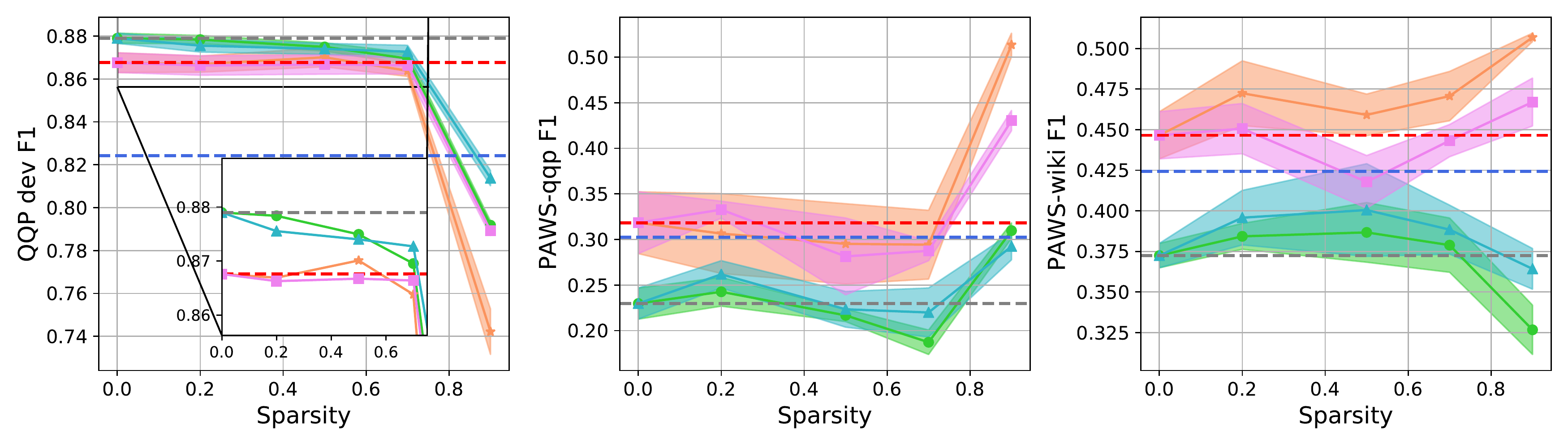}
        \includegraphics[width=1.0\linewidth]{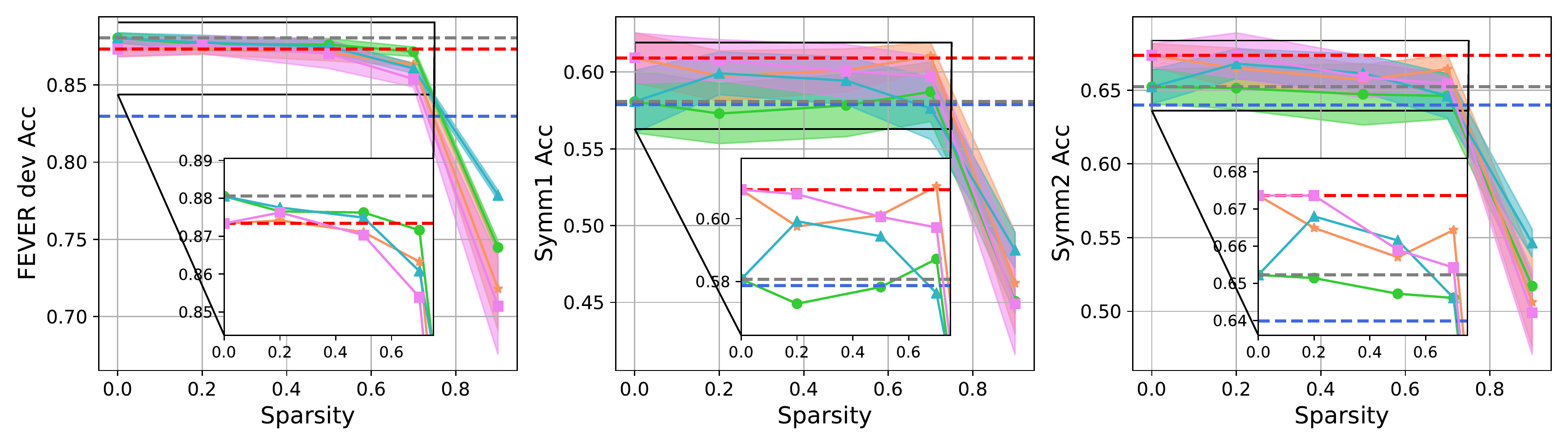}
    }
    \caption{Results of BERT subnetworks fine-tuned in isolation. ``ft'' is short for fine-tuning.}
    \label{fig:before_ft}
\end{figure}
%%%%%%%%%%%%%%%%%%%%%%%%%%%%%%%%%%%%%%%%%%%%%%

\subsection{BERT Subnetworks Fine-tuned in Isolation}
\label{sec:before_ft}
\subsubsection{Problem Formulation and Experimental Setups}
Given the pre-trained BERT $f(\boldsymbol{\theta}_{pt})$, a subnetwork $f(\mathbf{m} \odot \boldsymbol{\theta}_{pt})$ is obtained before downstream fine-tuning. The goal is to maximize the performance of the fine-tuned subnetwork $\mathcal{A}_{\mathcal{L}_1}(f(\mathbf{m} \odot \boldsymbol{\theta}_{pt}))$:
\begin{equation}
\label{eq:before_ft}
\max _{\mathbf{m}}\left(\mathcal{E}_{\text{ID}}\left(\mathcal{A}_{\mathcal{L}_1}(f(\mathbf{m} \odot \boldsymbol{\theta}_{pt}))\right) + \mathcal{E}_{\text{OOD}}\left(\mathcal{A}_{\mathcal{L}_1}(f(\mathbf{m} \odot \boldsymbol{\theta}_{pt}))\right) \right) ,
\text { s.t. } \frac{\|\mathbf{m}\|_{0}}{|\boldsymbol{\theta}_{pr}|} =(1-s)
\end{equation}
Following the LTH \cite{LTH}, we solve this problem using the train-prune-rewind pipeline. For IMP, the procedure is described in Section \ref{sec:magnitude_pruning} and $\mathbf{m} = \mathcal{M}(\mathcal{P}^{\text{imp-rw}}_{\mathcal{L}_2}(f(\boldsymbol{\theta}_{pt})))$. For mask training, the subnetwork structure is learned from $f(\boldsymbol{\theta}_{ft})$ (same as the previous section) and $\mathbf{m} = \mathcal{M}(\mathcal{P}^{\text{mask}}_{\mathcal{L}_2}(f(\boldsymbol{\theta}_{ft})))$.

We employ CE and PoE loss for model fine-tuning (i.e., $\mathcal{L}_1 \in \{\mathcal{L}_{\text{std}}, \mathcal{L}_{\text{poe}}\}$). Since we have shown that using the debiasing loss in pruning is conducive, the CE loss is not considered (i.e., $\mathcal{L}_2 = \mathcal{L}_{\text{poe}}$).

\subsubsection{Results}
The results of subnetworks fine-tuned in isolation are presented in Fig. \ref{fig:before_ft}. It can be found that: (1) For standard CE fine-tuning, the ``mask train (poe)'' subnetworks are superior to ``full bert (std)'' on the OOD test data, i.e., the subnetworks are less susceptible to the dataset bias during training. (2) In terms of the PoE-based fine-tuning, the ``imp (poe)'' and ``mask train (poe)'' subnetworks are generally comparable to ``full bert (poe)''. (3) For most of the subnetworks, ``poe ft'' clearly outperforms ``std ft'' in the OOD setting, which suggests that it is important to use the debiasing method in fine-tuning, even if the BERT subnetwork structure has already encoded some unbiased information.

Moreover, based on (1) and (2), we can extend the LTH on BERT \cite{BERT-LT,AllTicketWin,SuperTickets,TAMT}: \textbf{The pre-trained BERT contains SRNets that can be fine-tuned in isolation, using either standard or debiasing method, and match or even outperform the full model in both the ID and OOD evaluations.}

\subsection{BERT Subnetworks Without Fine-tuning}
\label{sec:no_ft}
\subsubsection{Problem Formulation and Experimental Setups}
This setup aims at finding a subnetwork $f(\mathbf{m} \odot \boldsymbol{\theta}_{pt})$ inside the pre-trained BERT, which can be directly employed to a task. The problem is formulated as:
\begin{equation}
\label{eq:no_ft}
\max _{\mathbf{m}}\left(\mathcal{E}_{\text{ID}}\left(f(\mathbf{m} \odot \boldsymbol{\theta}_{pt})\right) + \mathcal{E}_{\text{OOD}}\left(f(\mathbf{m} \odot \boldsymbol{\theta}_{pt})\right) \right) ,
\text { s.t. } \frac{\|\mathbf{m}\|_{0}}{|\boldsymbol{\theta}_{pr}|} =(1-s)
\end{equation}
Following \cite{Masking}, we fix the pre-trained parameters $\boldsymbol{\theta}_{pt}$ and optimize the mask variables $\mathbf{m}$. This process can be represented as $\mathcal{P}^{\text{mask}}_{\mathcal{L}}(f(\boldsymbol{\theta}_{pt}))$, where $\mathcal{L} \in \{\mathcal{L}_{\text{std}}, \mathcal{L}_{\text{poe}}\}$.

%%%%%%%%%%%%%%%%%%%%%%%%%%%%%%%%%%%%%%%%%%%%%%
\begin{figure}
    \parbox{0.25\linewidth}{
        \centering
        \includegraphics[width=1.0\linewidth]{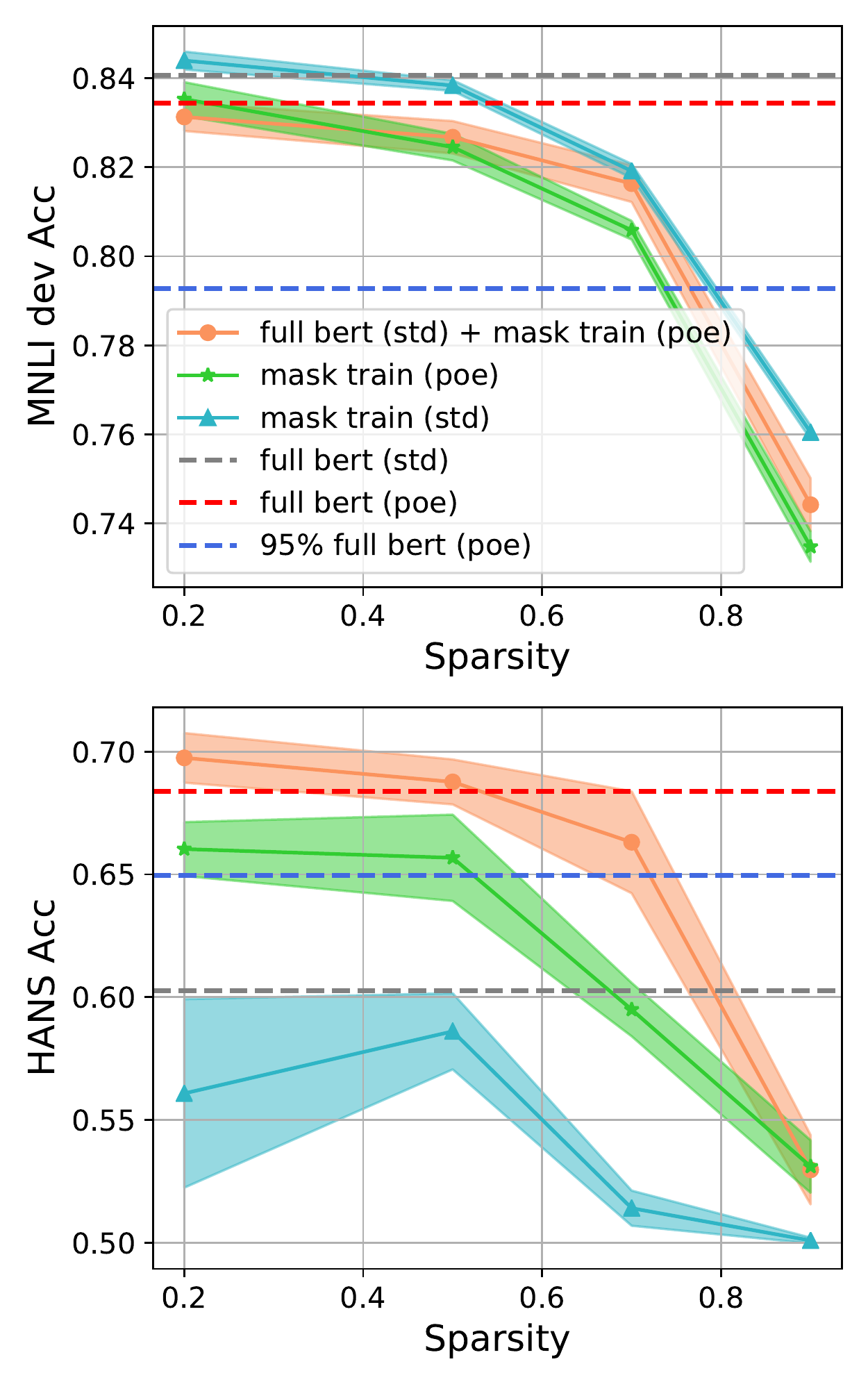}
    }
    \hfill
    \parbox{0.73\linewidth}{
        \centering
        \includegraphics[width=1.0\linewidth]{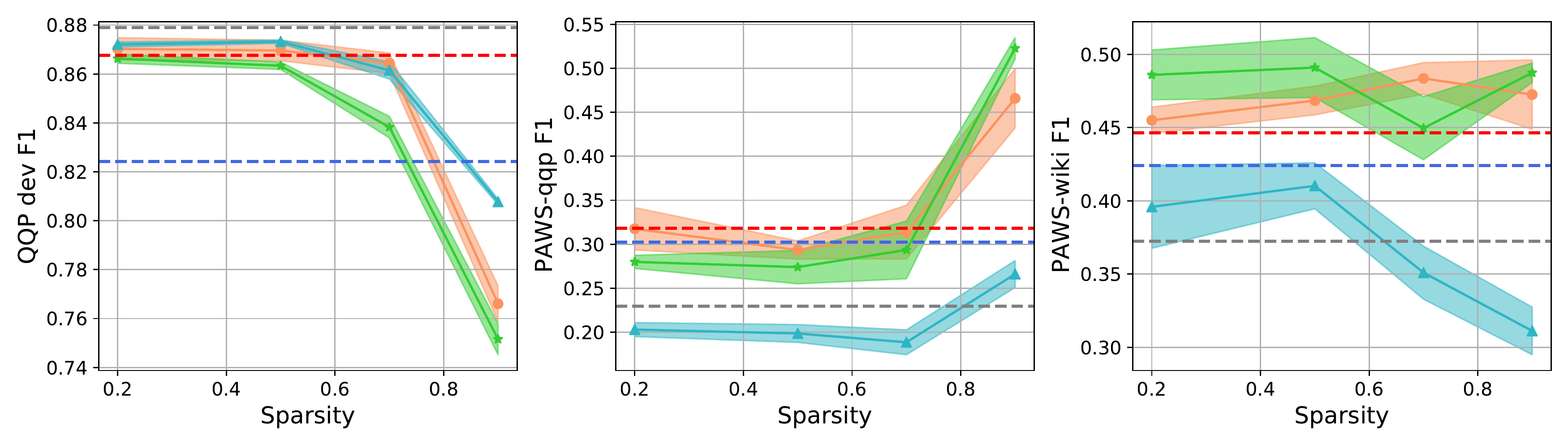}
        \includegraphics[width=1.0\linewidth]{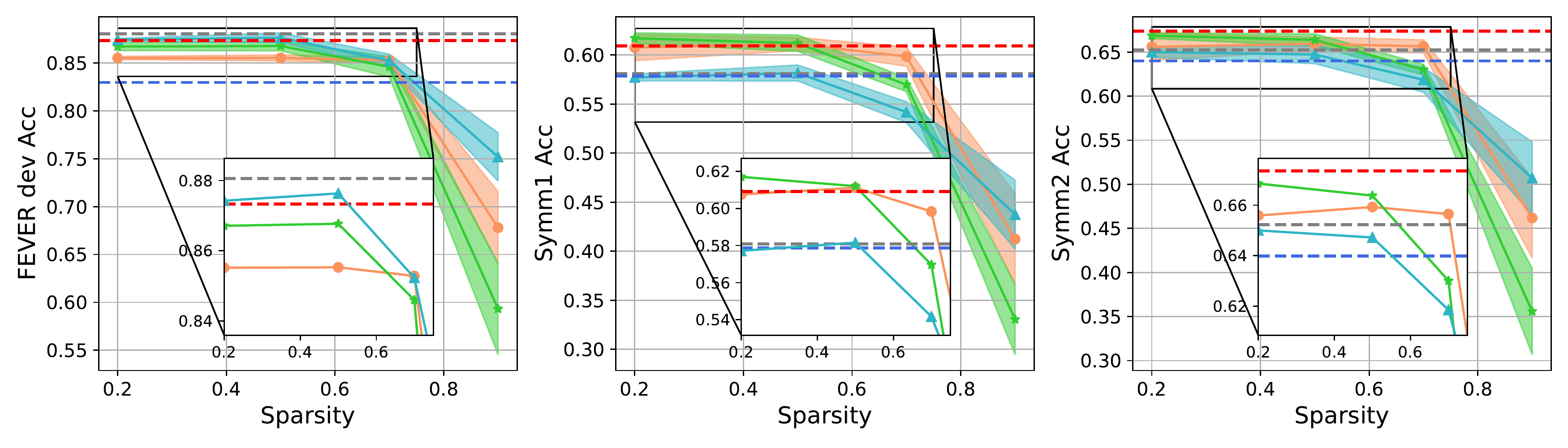}
    }
    \caption{Results of BERT subnetworks without fine-tuning. Results of the ``mask train (poe)'' subnetworks from Fig. \ref{fig:after_std_ft} (the orange line) are also reported for reference.}
    \label{fig:no_ft}
\end{figure}
%%%%%%%%%%%%%%%%%%%%%%%%%%%%%%%%%%%%%%%%%%%%%%
\subsubsection{Results}
\label{sec:subnet_wo_ft_results}
As we can see in Fig. \ref{fig:no_ft}: (1) With CE-based mask training, the identified subnetworks (under $50\%$ sparsity) in pre-trained BERT are competitive with the CE fine-tuned full BERT. (2) Similarly, using PoE-based mask training, the subnetworks under $50\%$ sparsity are comparable to the PoE fine-tuned full BERT, which demonstrates that SRNets for a particular downstream task already exist in the pre-trained BERT. (3) ``mask train (poe)'' subnetworks in pre-trained BERT can even match the subnetworks found in the fine-tuned BERT (the orange lines) in some cases (e.g., on PAWS and on FEVER under $50\%$ sparsity). Nonetheless, the latter exhibits a better overall performance.

\begin{figure*}[t]
    \parbox{0.49\linewidth}{
       \centering
        \includegraphics[width=1.\linewidth]{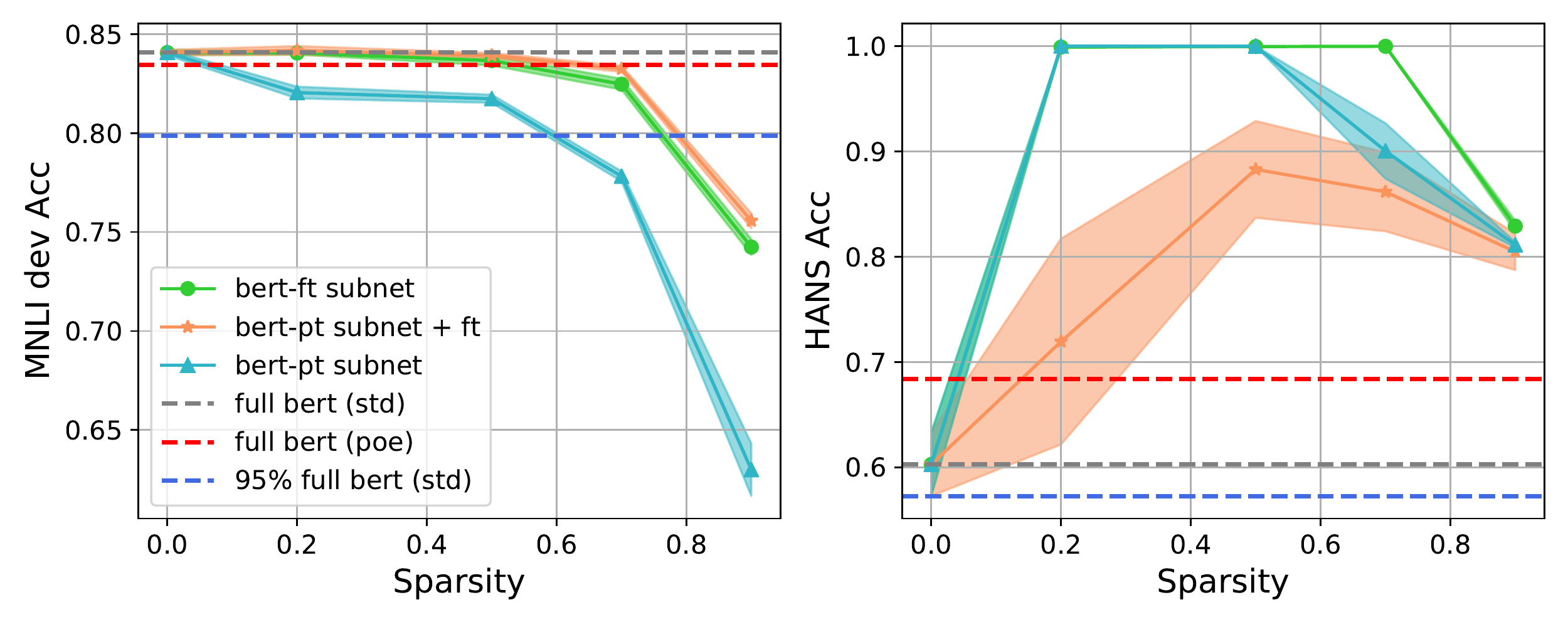}
        \caption{NLI results of BERT subnetworks found using the OOD information. Results of the other two tasks can be found in Appendix \ref{appendix-c2}.}
        \label{fig:oracle_mnli}
    }
    \hfill
    \parbox{0.49\linewidth}{
       \centering
        \includegraphics[width=1.\linewidth]{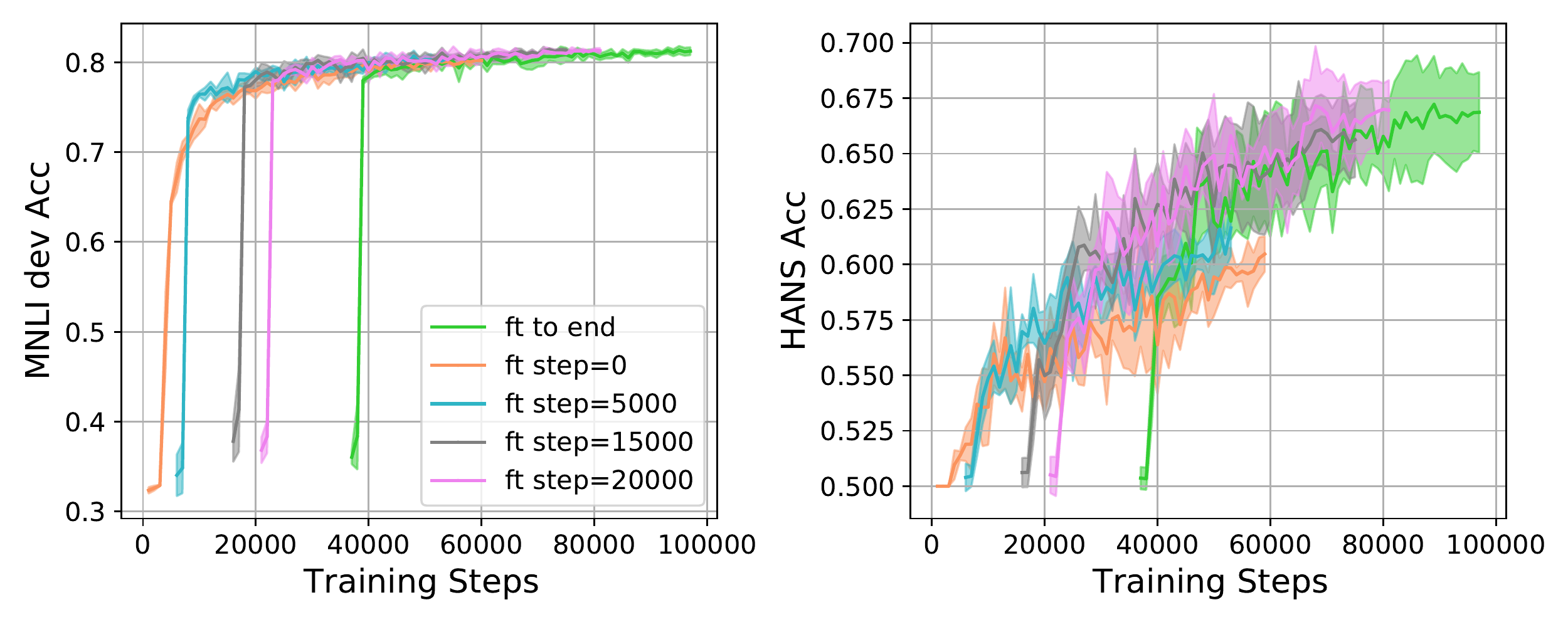}
        \caption{NLI mask training curves ($70\%$ sparse), starting from BERT fine-tuned for varied steps. Appendix \ref{appendix-c3} shows results of the other two tasks.}
        \label{fig:prune_timing_mnli}
    }
\end{figure*}

\subsection{Sparse and Unbiased BERT Subnetworks}
\subsubsection{Problem Formulation and Experimental Setups}
To explore the upper bound of BERT subnetworks in terms of OOD generalization, we include the OOD training data in mask training, and use the OOD test sets for evaluation. Like the previous sections, we investigate three pruning and fine-tuning paradigms, as formulated by Eq. \ref{eq:after_ft}, \ref{eq:before_ft} and \ref{eq:no_ft} respectively. We only consider the standard CE for subnetwork and full BERT fine-tuning, which is more vulnerable to the dataset bias. Appendix \ref{appendix-b3.3} summarizes the detailed experimental setups.

\subsubsection{Results}
From Fig. \ref{fig:oracle_mnli} we can observe that: (1) The subnetworks from fine-tuned BERT (``bert-ft subnet'') at $20\% \sim 70\%$ sparsity achieve nearly $100\%$ accuracy on HANS, and their ID performance is also close to the full BERT. (2) The subnetworks in the pre-trained BERT (``bert-pt subnet'') also have very high OOD accuracy, while they perform worse than ``bert-ft subnet'' in the ID setting. (3) ``bert-pt subnet + ft'' subnetworks, which are fine-tuned in isolation with CE loss, exhibits the best ID performance, and the poorest OOD performance. However, compared to the full BERT, these subnetworks still rely much less on the dataset bias, reaching nearly $90\%$ HANS accuracy at $50\%$ sparsity. Jointly, these results show that there consistently exist BERT subnetworks that are almost unbiased towards the MNLI training set bias, under the three kinds of pruning and fine-tuning paradigms.

%%%%%%%%%%%%%%%%%%%%%%%%%%%%%%%%%%%%%%%%%%%%%%
\begin{figure*}[t]
\centering
\includegraphics[width=0.8\linewidth]{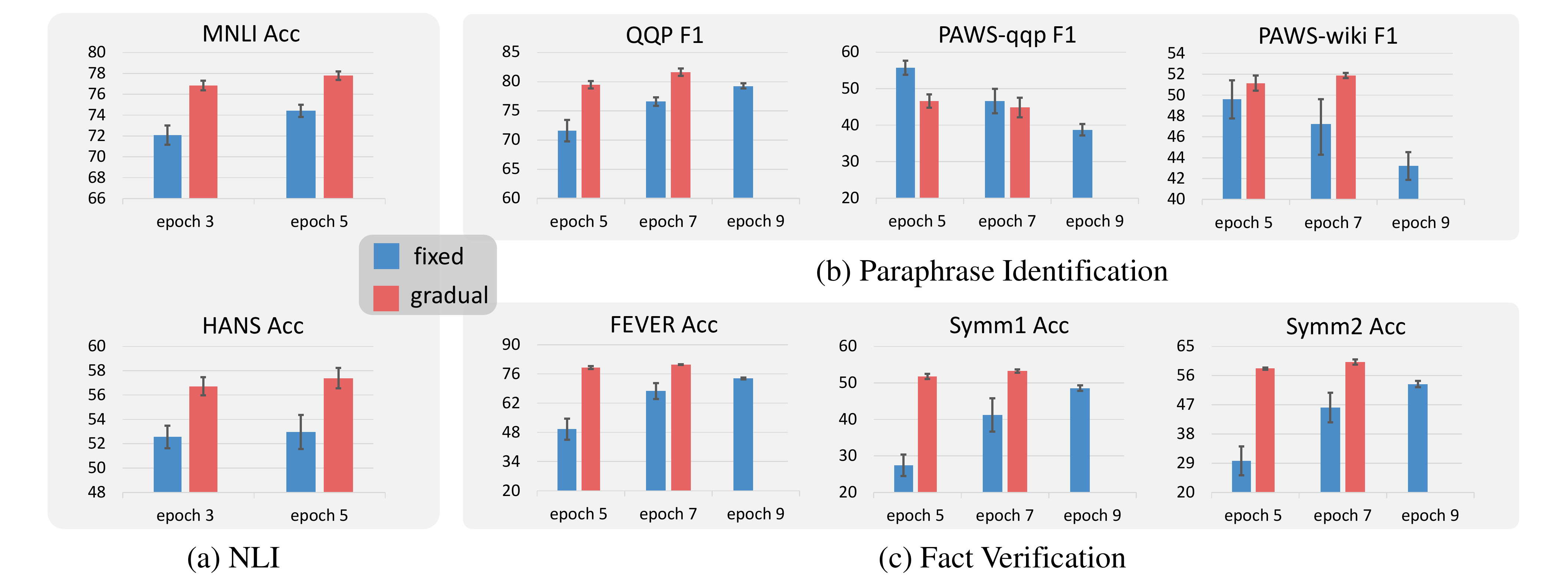}
\caption{Comparison between fixed sparsity and gradual sparsity increase for mask training with the standard fine-tuned full BERT. The subnetworks are at $90\%$ sparsity.}
% The horizontal axis represents the number of training epochs. 
\label{fig:graduL_sparsity_increase}
\end{figure*}
%%%%%%%%%%%%%%%%%%%%%%%%%%%%%%%%%%%%%%%%%%%%%%

\section{Refining the SRNets Searching Process}
In this section, we study how to further improve the SRNets searching process based on mask training, which generally performs better than IMP, as shown in Section \ref{sec:after_ft} and Section \ref{sec:before_ft}.

\subsection{The Timing to Start Searching SRNets}
\label{sec:timing}
Compared with searching subnetworks from the fine-tuned BERT, directly searching from the pre-trained BERT is more efficient in that it dispenses with fine-tuning the full model. However, the former has a better overall performance, as we have shown in Section \ref{sec:no_ft}. This induces a question: \textbf{At which point of the BERT fine-tuning process, can we find subnetworks comparable to those found after the end of fine-tuning using mask training?} To answer this question, we perform mask training on the model checkpoints $f(\boldsymbol{\theta}_t) = \mathcal{A}^{t}_{\mathcal{L}_{\text{std}}}(f(\boldsymbol{\theta}_{pt}))$ from different steps $t$ of BERT fine-tuning.

Fig. \ref{fig:prune_timing_mnli} shows the mask training curves, which start from different $f(\boldsymbol{\theta}_t)$. We can see that ``ft step=0'' converges slower and to a worse final accuracy, as compared with ``ft to end'', especially on the HANS dataset. However, with 20,000 steps of full BERT fine-tuning, which is roughly $55\%$ of the ``ft to end'', the mask training performance is very competitive. This suggests that the total training cost of SRNet searching can be reduced, by a large amount, in the full model training stage.

To actually reduce the training cost, we need to predict the exact timing to start mask training. This is intractable without information of all the training curves in Fig. \ref{fig:prune_timing_mnli}. A feasible solution is adopting the idea of early-stopping (see Appendix \ref{appendix-e1} for detailed discussions). However, accurately predicting the optimal timing (with the least amount of fine-tuning and comparable subnetwork performance to fully fine-tuning) is indeed difficult and we invite follow-up studies to investigate this question. 

\subsection{SRNets at High Sparsity}
As the results of Section \ref{sec:srnets} demonstrate, there is a sharp decline of the subnetworks' performance from $70\% \sim 90\%$ sparsity. We conjecture that this is because directly initializing mask training to $90\%$ reduces the model's capacity too drastically, and thus causes some difficulties in optimization. Therefore, we gradually increase the sparsity from $70\% \sim 90\%$ during mask training, using the cubic sparsity schedule \cite{ToPruneOrNotToPrune} (see Appendix \ref{appendix-c4} for ablation studies).
% using the soft variant of initialization (see Section \ref{sec:mask_train}) 
Fig. \ref{fig:graduL_sparsity_increase} compares the fixed sparsity used in the previous sections and the gradual sparsity increase, across varied mask training epochs. We find that while simply extending the training process is conducive, gradual sparsity increase achieves better results. In particular, ``gradual'' outperforms ``fixed'' with lower training cost on all the three tasks, except for the PAWS dataset, A similar phenomenon is explained in Section \ref{sec:result_after_ft}.

\section{Conclusions and Limitations}
\label{sec:conclusion}
In this paper, we investigate whether sparsity and robustness to dataset bias can be achieved simultaneously for PLM subnetworks. Through extensive experiments, we demonstrate that BERT indeed contains sparse and robust subnetworks (SRNets) across a variety of NLU tasks and training and pruning setups. We further use the OOD information to reveal that there exist sparse and almost unbiased BERT subnetworks. Finally, we present analysis and solutions to refine the SRNet searching process in terms of subnetwork performance and searching efficiency.

The limitations of this work is twofold. First, we focus on BERT-like PLMs and NLU tasks, while dataset biases are also common in other scenarios. For example, gender and racial biases exist in dialogue generation systems \cite{QueensArePowerful} and PLMs \cite{AutoDebias}. In the future work, we would like to extend our exploration to other types of PLMs and NLP tasks (see Appendix \ref{appendix-e2} for a discussion). Second, as we discussed in Section \ref{sec:timing}, our analysis on “the timing to start searching SRNets” mainly serves as a proof-of-concept, and actually reducing the training cost requires predicting the exact timing.

\begin{ack}
This work was supported by National Natural Science Foundation of China (61976207 and 61906187).
\end{ack}

\bibliography{neurips_2022}
\bibliographystyle{abbrvnat}

%%%%%%%%%%%%%%%%%%%%%%%%%%%%%%%%%%%%%%%%%%%%%%%%%%%%%%%%%%%%
\section*{Checklist}

%%% BEGIN INSTRUCTIONS %%%
% The checklist follows the references.  Please
% read the checklist guidelines carefully for information on how to answer these
% questions.  For each question, change the default \answerTODO{} to \answerYes{},
% \answerNo{}, or \answerNA{}.  You are strongly encouraged to include a {\bf
% justification to your answer}, either by referencing the appropriate section of
% your paper or providing a brief inline description.  For example:
% \begin{itemize}
%   \item Did you include the license to the code and datasets? \answerYes{See Section~\ref{gen_inst}.}
%   \item Did you include the license to the code and datasets? \answerNo{The code and the data are proprietary.}
%   \item Did you include the license to the code and datasets? \answerNA{}
% \end{itemize}
% Please do not modify the questions and only use the provided macros for your
% answers.  Note that the Checklist section does not count towards the page
% limit.  In your paper, please delete this instructions block and only keep the
% Checklist section heading above along with the questions/answers below.
%%% END INSTRUCTIONS %%%

\begin{enumerate}

\item For all authors...
\begin{enumerate}
  \item Do the main claims made in the abstract and introduction accurately reflect the paper's contributions and scope?
    \answerYes{}
  \item Did you describe the limitations of your work?
    \answerYes{See Section \ref{sec:conclusion}.}
  \item Did you discuss any potential negative societal impacts of your work?
    \answerNo{Currently, we think there are no apparent negative societal impacts related to our work.}
  \item Have you read the ethics review guidelines and ensured that your paper conforms to them?
    \answerYes{}
\end{enumerate}

\item If you are including theoretical results...
\begin{enumerate}
  \item Did you state the full set of assumptions of all theoretical results?
    \answerNA{}
        \item Did you include complete proofs of all theoretical results?
    \answerNA{}
\end{enumerate}

\item If you ran experiments...
\begin{enumerate}
  \item Did you include the code, data, and instructions needed to reproduce the main experimental results (either in the supplemental material or as a URL)?
    \answerNo{We will release the codes and reproduction instructions upon publication.}
  \item Did you specify all the training details (e.g., data splits, hyperparameters, how they were chosen)?
    \answerYes{See Section \ref{sec:experi_setups} and Appendix \ref{appendix-b}.}
        \item Did you report error bars (e.g., with respect to the random seed after running experiments multiple times)?
    \answerYes{See all the figures of our experiments.}
        \item Did you include the total amount of compute and the type of resources used (e.g., type of GPUs, internal cluster, or cloud provider)?
    \answerYes{See Appendix \ref{appendix-b}.}
\end{enumerate}

\item If you are using existing assets (e.g., code, data, models) or curating/releasing new assets...
\begin{enumerate}
  \item If your work uses existing assets, did you cite the creators?
    \answerYes{See Section \ref{sec:experi_setups}.}
  \item Did you mention the license of the assets?
    \answerYes{Licenses of some dataset we used are mentioned in Section \ref{sec:experi_setups}. However, for the other datasets, we were unable to find the licenses.}
  \item Did you include any new assets either in the supplemental material or as a URL?
    \answerNo{}
  \item Did you discuss whether and how consent was obtained from people whose data you're using/curating?
    \answerNA{}
  \item Did you discuss whether the data you are using/curating contains personally identifiable information or offensive content?
    \answerNA{}
\end{enumerate}

\item If you used crowdsourcing or conducted research with human subjects...
\begin{enumerate}
  \item Did you include the full text of instructions given to participants and screenshots, if applicable?
    \answerNA{}
  \item Did you describe any potential participant risks, with links to Institutional Review Board (IRB) approvals, if applicable?
    \answerNA{}
  \item Did you include the estimated hourly wage paid to participants and the total amount spent on participant compensation?
    \answerNA{}
\end{enumerate}

\end{enumerate}

%%%%%%%%%%%%%%%%%%%%%%%%%%%%%%%%%%%%%%%%%%%%%%%%%%%%%%%%%%%%

\appendix
\section{More Information of Pruning and Debiasing Methods}
\subsection{Pruning Methods}
\label{appendix-a1}
\subsubsection{Iterative Magnitude Pruning}
\label{appendix-a1.1}
Algo. \ref{alg:imp} summarizes our implementation of IMP and IMP with weight rewinding. In practice, we set the per time pruning ratio $\Delta{s}=10\%$ and the pruning interval $\Delta{t}=0.1 \cdot t_{\text{max}}$.

%%%%%%%%%%%%%%%%%%%%%%%%%%%%%%%%%%%%%%%%%%%%%%
\begin{algorithm}[htb] \label{alg:imp}
\SetAlgoLined
\KwIn{PLM $f(\boldsymbol{\theta}_{0})$ w. $\boldsymbol{\theta}_0 = \boldsymbol{\theta}_{ft}$, maximum training steps $t_{\text{max}}$, pruning interval $\Delta{t}$, per time pruning ratio $\Delta{s}$, target sparsity level $s=k \cdot \Delta{s}$ ($k \in \{1, 2, \cdots\}$), pruning method $p \in \{\text{imp}, \text{imp-rw}\}$}
\KwOut{Pruned subentwork $f(\mathbf{m}\odot\boldsymbol{\theta}^{'}_{ft})$}
Initialize the pruning mask $\mathbf{m}=1^{|\boldsymbol{\theta}_0|}$ and the number of pruning $n=0$\\
\While{t $<$ $t_{\text{max}}$}{
\If{(t mod $\Delta{t}$) == $0$}{
    \textcolor{gray}{\texttt{\# For imp, return the subnetwork after some further training}}\\
    \If{$n \cdot \Delta{s}$ == $s$ and $p$==$\operatorname{imp}$}{
        \textbf{return} $f(\mathbf{m} \odot \boldsymbol{\theta}_t)$
    }
    Prune $\Delta{s} \cdot |\boldsymbol{\theta}_{0}|$ from the remaining parameters $\mathbf{m} \odot \boldsymbol{\theta}_t$ based on the magnitudes, and update $\mathbf{m}$ accordingly \\
    $n \leftarrow n+1$ \\
    \textcolor{gray}{\texttt{\# For imp-rw, return the subnetwork directly after pruning}}\\
    \If{$n \cdot \Delta{s}$ == $s$ and $p$==$\operatorname{imp-rw}$}{
        \textbf{return} $f(\mathbf{m} \odot \boldsymbol{\theta}_0)$
    }
}
Update the remaining model parameters $\mathbf{m} \odot \boldsymbol{\theta}_t$ via AdamW \cite{AdamW}; \\
}
\caption{Iterative Magnitude Pruning (+ weight rewinding)}
\end{algorithm}
%%%%%%%%%%%%%%%%%%%%%%%%%%%%%%%%%%%%%%%%%%%%%%

\subsubsection{Mask Training}
\label{appendix-a1.2}
As we described in Section \ref{sec:mask_train} of the main paper, we realize mask training via binarization in forward pass and gradient estimation in backward pass. Following \cite{HowFine,TAMT}, we adopt a magnitude-based strategy to initialize the real-valued masks. Specially, we consider two variants: The first one (hard variant) identifies the weights in matrix $\mathbf{W}$ with the smallest magnitudes, and sets the corresponding elements in $\hat{\mathbf{m}}$ to zero, and the remaining elements to a fixed value:
\begin{equation}
\label{eq:omp_init_hard}
\hat{\mathbf{m}}_{i, j}= \begin{cases} 0 & \text { if } \mathbf{W}_{i, j} \in \operatorname{Min}_{s}(\operatorname{abs}(\mathbf{W})) \\ \alpha \times \phi & \text { otherwise }\end{cases}
\end{equation}
where $\operatorname{Min}_{s}(\operatorname{abs}(\mathbf{W}))$ extracts the weights with the lowest absolute value, according to sparsity level $s$. $\alpha \geq 1$ is a hyper-parameter. The second one (soft variant) directly utilizes the absolute values of the weights for mask initialization:
\begin{equation}
    \label{eq:omp_init_soft}
    \hat{\mathbf{m}}_{i, j} = \operatorname{abs}(\mathbf{W}_{i, j})
\end{equation}

To control the sparsity of the model, the threshold $\phi$ is adjusted dynamically at a frequency of $\Delta{t}_{\phi}$ training steps. In practice, we control the sparsity in a local way, i.e., all the weight matrices $\mathbf{W} \in \boldsymbol{\theta}_{pr}$ should satisfy the same sparsity constraint $s$. Algo. \ref{alg:mask_train} summarizes the entire process of mask training.

%%%%%%%%%%%%%%%%%%%%%%%%%%%%%%%%%%%%%%%%%%%%%%
\begin{algorithm}[htb] \label{alg:mask_train}
\SetAlgoLined
\KwIn{PLM $f(\boldsymbol{\theta}_{0})$ w. $\boldsymbol{\theta}_0 \in \{\boldsymbol{\theta}_{pt}, \boldsymbol{\theta}_{ft}\}$, maximum training steps $t_{\text{max}}$, frequency $\Delta{t}_{\phi}$, target sparsity level $s$, threshold $\phi$, hyper-parameter $\alpha$, initialization method $init \in \{\text{hard}, \text{soft}\}$}

\KwOut{Pruned subentwork $f(\mathbf{m} \odot \boldsymbol{\theta}_0)$}

\eIf{init == $\operatorname{hard}$}{
    Initialize the real-valued mask $\hat{\mathbf{m}}$ according to Eq. \ref{eq:omp_init_hard} \\
    Set threshold $\phi = 0.01$
}{
    Initialize the real-valued mask $\hat{\mathbf{m}}$ according to Eq. \ref{eq:omp_init_soft} \\
    Set threshold $\phi$ according to the sparsity constraint
}

\While{t $<$ $t_{\text{max}}$}{
    Get a mini-batch of $B$ examples $\{(\mathbf{x}_b, y_b)\}^{B}_{b=1}$\\
    Forward pass through binarization:\\
    \qquad $\mathcal{L}(f(\mathbf{x}_b, \mathbf{m} \odot \boldsymbol{\theta}_0), y_b)$,
    \qquad where $\mathbf{m}_{i, j}= \begin{cases}1 & \text { if } \hat{\mathbf{m}}_{i, j} \geq \phi \\ 0 & \text { otherwise }\end{cases}$ \\
    
    Backward pass through gradient estimation:\\
    \qquad $\hat{\mathbf{m}} \leftarrow \hat{\mathbf{m}}-\eta \frac{\partial \mathcal{L}}{\partial \mathbf{m}}$ \\
    
    \If{(t mod $\Delta{t}_{\phi}$) == $0$}{
        Update the threshold $\phi$ to satisfy the sparsity constraint
    }
}
\textbf{return} $f(\mathbf{m} \odot \boldsymbol{\theta}_0)$ \\
\caption{Mask Training}
\end{algorithm}
%%%%%%%%%%%%%%%%%%%%%%%%%%%%%%%%%%%%%%%%%%%%%%

\subsection{Debiasing Methods}
\label{appendix-a2}
We have introduced the PoE method in Section \ref{sec:debiasing}. Here we provide descriptions of the other two debiasing methods, i.e., example reweighting and confidence regularization.

\textbf{Example Reweighting} directly assigns an importance weight to the standard CE training loss, according to the bias degree $\beta$:
\begin{equation}
\mathcal{L}_{\text{reweight}}=-\left(1-\beta\right) \mathbf{y} \cdot \log \mathbf{p}_m
\end{equation}

\textbf{Confidence Regularization} is based on knowledge distillation \cite{KD}. It involves a teacher model trained with the standard CE loss. The teacher model's prediction $\mathbf{p}_t$ is used as a supervision signal to train the main model. To account for the bias degree of training examples, $\mathbf{p}_t$ is smoothed using a scaling function $\mathrm{S}\left(\mathbf{p}_t, \beta \right)$, and the final loss is computed as:
\begin{equation}
\begin{aligned}
    &\mathcal{L}_{\text{confreg}}=-\mathrm{S}\left(\mathbf{p}_t, \beta \right) \cdot \log \mathbf{p}_m \\&
    \mathrm{S}\left(\mathbf{p}_t, \beta \right)=\frac{(\mathbf{p}^{j}_t)^{\left(1-\beta\right)}}{\sum_{k=1}^{K} (\mathbf{p}^{k}_t)^{\left(1-\beta\right)}}
\end{aligned}
\end{equation}

\section{More Experimental Setups}
\label{appendix-b}
\subsection{Datasets and Evaluations}
\label{appendix-b1}

%%%%%%%%%%%%%%%%%%%%%%%%%%%%%%%%%%
\begin{table*}[t]
\centering
\caption{The number of examples in different dataset splits. The splits used for evaluation are highlighted with red color. The dev set for MNLI is MNLI-m.}
\label{tab:data_num}

\resizebox{1\hsize}{!}{$
\begin{tabular}{@{}l c c c c c c c c@{}}
\toprule
\multirow{2}{*}{} & \multicolumn{2}{c}{NLI} & \multicolumn{3}{c}{Paraphrase Identification} & \multicolumn{3}{c}{Fact Verification}\\
\cmidrule(r){2-3} \cmidrule(r){4-6} \cmidrule(r){7-9}
&MNLI  &HANS   &QQP  &PAWS-qqp     &PAWS-wiki     &FEVER    &FEVER-Symm1     &FEVER-Symm2   \\ \midrule
Train   &392,702  &30,000   &363,849  &11,988   &49,401   &242,911  &-    &-       \\
Dev     &\color{red}{9,815}    &\color{red}{30,000}   &\color{red}{40,432}   &\color{red}{677}      &8,000    &\color{red}{16,664}   &-    &708      \\
Test    &-        &-        &-        &-        &\color{red}{8,000}    &-        &\color{red}{717}  &\color{red}{712}       \\
\bottomrule
\end{tabular}
$}
\end{table*}
%%%%%%%%%%%%%%%%%%%%%%%%%%%%%%%%%%%%%%%%%%%
%%%%%%%%%%%%%%%%%%%%%%%%%%%%%%%%%%%%%%%%%%%%%%
\begin{table}[t]
    \caption{Data distribution over classes. The meaning of the abbreviations are: ent (entailment), cont (contradiction), dulp (duplicate), supp (support), not-info (not-enough-info). ``Eval'' represents the dataset split used for evaluation, as described in Tab. \ref{tab:data_num}}
    \label{tab:data_class}
    \parbox{0.25\linewidth}{
        \centering
        \resizebox{1.0\hsize}{!}{$
            \begin{tabular}{@{}l l c c c @{}}
                \toprule
                &                        &MNLI   &HANS     \\\midrule
                \multirow{3}{*}{Train}
                     &ent         &33.3\%     &50\%      \\
                     &cont     &33.3\%     &50\%        \\
                     &neutral        &33.3\%     &0\%   \\\midrule
                \multirow{3}{*}{Eval}
                     &ent         &35.4\%     &50\%      \\
                     &cont     &32.7\%     &50\%        \\
                     &neutral        &31.8\%     &0\%    \\
                \bottomrule
                \end{tabular}
        $}
    }
    \hfill
    \parbox{0.36\linewidth}{
        \centering
        \resizebox{1.0\hsize}{!}{$
            \begin{tabular}{@{}l l c c c c @{}}
                \toprule
                &           &QQP   &$\text{PAWS}_{qqp}$ &$\text{PAWS}_{wiki}$     \\\midrule
                \multirow{3}{*}{Train}
                     &dulp        &36.9\%   &31.5\%   &44.2\%  \\
                     &non-dulp    &63.1\%   &68.5\%   &55.8\%  \\
                     \\\midrule
                \multirow{3}{*}{Eval}
                     &dulp         &36.8\%   &28.2\%   &44.2\%  \\
                     &non-dulp     &63.2\%   &71.8\%   &55.8\%  \\
                     \\
                \bottomrule
                \end{tabular}
        $}
    }
    \hfill
    \parbox{0.36\linewidth}{
        \centering
        \resizebox{1.0\hsize}{!}{$
            \begin{tabular}{@{}l l c c c c @{}}
                \toprule
                &                       &FEVER   &Symm1    &Symm2     \\\midrule
                \multirow{3}{*}{Train}
                     &supp         &41.4\%     &- &- \\
                     &refute       &17.2\%     &- &- \\ 
                     &not-info      &41.4\%     &- &- \\\midrule
                \multirow{3}{*}{Eval}
                     &supp         &47.9\%    &52.9\%     &50\% \\
                     &refute       &52.1\%    &47.1\%     &50\% \\ 
                     &not-info      &0\%       &0\%        &0\% \\
                \bottomrule
                \end{tabular}
        $}
    }
\end{table}
%%%%%%%%%%%%%%%%%%%%%%%%%%%%%%%%%%%%%%%%%%%%%%

We utilize eight datasets from three NLU tasks. The statistics of different dataset splits are summarized in Tab. \ref{tab:data_num}. If one dataset has a test set, we use it for evaluation, and otherwise we report results on the dev set. For MNLI and QQP, since the official test server \footnote{\url{https://gluebenchmark.com/}} only allows two submissions a day, we instead evaluate on the dev sets, following \cite{BERT-LT,TAMT,MovementPruning}. For FEVER, we use the training and evaluation data processed by \cite{FeverSym} \footnote{\url{https://github.com/TalSchuster/FeverSymmetric}}.

Tab. \ref{tab:data_class} shows the distribution of examples over classes. We can see that the distributions of the QQP and $\text{PAWS}_{qqp}$ evaluation sets are imbalanced. Specially, in the OOD $\text{PAWS}_{qqp}$, where a biased model tends to predict most examples to the \textit{duplicate} class, simply classifying all examples as \textit{non-duplicate} can achieve substantial improvement in accuracy (from $28.2\%$ to $71.8\%$). To account for this, we use the F1 score to evaluate the performance on the three paraphrase identification datasets. Specifically, we calculate the weighted average of the F1 score of each class. However, the class imbalance may still affect the evaluation on PAWS (as we discussed in Section \ref{sec:result_after_ft}) and therefore the OOD improvement should be assessed by also considering the ID performance.

\subsection{Software and Computational Resources}
\label{appendix-b2}
We use two types of GPU, i.e., Nvidia V100 and TITAN RTX. All the experiments are run on a single GPU. Our codes are based on the Pytorch\footnote{\url{https://pytorch.org/}} and the huggingface transformers library\footnote{\url{https://github.com/huggingface/transformers}} \cite{huggingface}.

\subsection{Training Details}
\label{appendix-b3}
\subsubsection{Bias Model}
\label{appendix-b3.1}
As mentioned in Section \ref{sec:train_details}, we train the bias model with spurious features. For MNLI and QQP, we adopt the hand-crafted word overlapping features proposed by \cite{DontTakeEasyWay}, which includes: 
\begin{itemize}
    \item Whether all the hypothesis words also belong to the premise.
    \item Whether the hypothesis appears as a continuous subsequence in the premise.
    \item The percentage of the hypothesis words $\mathbf{w}^h = \{\mathbf{w}^h_{1}, \mathbf{w}^h_{2}, \cdots, \mathbf{w}^h_{|\mathbf{w}^h|}\}$ that appear in the premise $\mathbf{w}^p = \{\mathbf{w}^p_{1}, \mathbf{w}^p_{2}, \cdots, \mathbf{w}^p_{|\mathbf{w}^p|}\}$. Formally $\frac{|\mathbf{w}^h \cap \mathbf{w}^p|}{|\mathbf{w}^h|}$.
    \item The average of the maximum similarity between each hypothesis word and all the premise words: $\frac{1}{|\mathbf{w}^h|} \operatorname{sum}(\{\operatorname{max}(\{\operatorname{sim}(\mathbf{w}^p_{i}, \mathbf{w}^h_{j}) \vert \forall \mathbf{w}^p_{j} \in \mathbf{w}^p \}) \vert \forall \mathbf{w}^h_{i} \in \mathbf{w}^h\})$, where the similarity is computed based on the fastText word vectors \cite{fasttext} and the cosine distance.
    \item The minimum of the same similarities above: $\operatorname{min}(\{\operatorname{max}(\{\operatorname{sim}(\mathbf{w}^p_{i}, \mathbf{w}^h_{j}) \vert \forall \mathbf{w}^p_{j} \in \mathbf{w}^p \}) \vert \forall \mathbf{w}^h_{i} \in \mathbf{w}^h\}).$
\end{itemize}

For FEVER, we use the max-pooled word embeddings of the claim sentence, which are also based on the fastText word vectors.

%%%%%%%%%%%%%%%%%%%%%%%%%%%%%%%%%%
\begin{table*}[t]
\centering
\caption{Basic training hyper-parameters.}
\label{tab:base-hyperparam}

\resizebox{1. \hsize}{!}{$
\begin{tabular}{@{}l c c c c c c c @{}}
\toprule
 & \#Epoch & Learning Rate & Batch Size &  Max Length & Eval Interval & Eval Metric & Optimizer \\ \midrule
MNLI    &3 or 5    &5e-5    &32    &128      &1,000    &Acc    &AdamW         \\
QQP     &3    &2e-5    &32    &128      &1,000    &F1     &AdamW         \\
FEVER   &3    &2e-5    &32    &128      &500      &Acc    &AdamW          \\
\bottomrule
\end{tabular}
$}
\end{table*}
%%%%%%%%%%%%%%%%%%%%%%%%%%%%%%%%%%%%%%%%%%%
%%%%%%%%%%%%%%%%%%%%%%%%%%%%%%%%%%
\begin{table*}[t]
\centering
\caption{Basic hyper-parameters related to pruning methods. $t_{\text{max}}$ is the number of optimization steps by training \#Epoch epochs.}
\label{tab:base-hyperparam-pruning}

\resizebox{0.9 \hsize}{!}{$
\begin{tabular}{@{}l c c c c c c c @{}}
\toprule
\multirow{2}{*}{} & \multicolumn{5}{c}{Mask Training} & \multicolumn{2}{c}{IMP} \\
\cmidrule(r){2-6} \cmidrule(r){7-8}
&Mask Init  &Sparsity Schedule  &$\phi$  &$\alpha$   &$\Delta{t}_{\phi}$  &$\Delta{s}$  &$\Delta{t}$ \\ \midrule
&magnitude (hard)    &fixed to $s$    &0.01    &2      &equal to Eval Interval    &10\%    &0.1$\cdot t_{\text{max}}$         \\
\bottomrule
\end{tabular}
$}
\end{table*}
%%%%%%%%%%%%%%%%%%%%%%%%%%%%%%%%%%%%%%%%%%%
\subsubsection{Full BERT}
\label{appendix-b3.2}
The main training hyper-parameters are shown in Tab. \ref{tab:base-hyperparam}, which basically follow \cite{UnknownBias}. Most of the hyper-parameters are the same for different training strategies, except for the number of training epochs (\#Epoch) on MNLI. For the standard CE loss and example reweighting, the model is trained for 3 epochs. For PoE and confidence regularization, the model is trained for 5 epochs.

\subsubsection{Mask Training and IMP}
\label{appendix-b3.3}
Mask training and IMP basically use the same set of hyper-parameters as full BERT, except for longer training. The number of training epochs for mask training and IMP is 5 on MNLI, and 7 on QQP and FEVER. The hyper-parameters specific to mask training or IMP are summarized in Tab. \ref{tab:base-hyperparam-pruning}. Unless otherwise specified, we adopt the hard-variant of mask initialization (Eq. \ref{eq:omp_init_hard}) and fix the subnetwork sparsity to target sparsity $s$ throughout the process of mask training. Some special experimental setups are described as follows:

\paragraph{Subnetworks from Fine-tuned BERT}
When we search for subnetworks at low sparsity (e.g., 20\%) from a fine-tuned BERT, we find that mask training (with debiasing loss) stably improves the OOD performance, while the ID performance peaks at an early point of training and then slightly drops and recovers later. Therefore, the ID performance favors the early checkpoints, which are not good at the OOD generalization. To address this problem, we select the best checkpoint after $0.7 \cdot t_{\text{max}}$ of training, but still according to the performance on the ID dev set. This strategy is only adopted for mask training on fine-tuned BERT (for all sparsity levels), and in other cases we select the best checkpoint across training based on ID performance.

\paragraph{BERT Subnetworks Fine-tuned in Isolation}
When fine-tuning the searched subnetworks (with their weights rewound to pre-trained values) in isolation, we use the same set of hyper-parameters as full BERT fine-tuning.

\paragraph{Sparse and Unbiased BERT Subnetworks}
The OOD data is used in this setup. Specifically, we utilize the training data of HANS and PAWS for NLI and paraphrase identification respectively. In terms of the FEVER-Symmetric dataset, which does not provide a training set (see Tab. \ref{tab:data_num}), we use the dev set of FEVER-Symm2 and copy the data 10 times to construct the OOD training data. The OOD and ID training data are then combined to form the final training set. Note that the evaluation sets are the same as the other setups, and \textbf{NO} test data is used in mask training.

\paragraph{Gradual Sparsity Increase} We mainly experiment with the gradual sparsity increase schedule for subnetworks at 90\% sparsity. Concretely, we increase the sparsity from 70\% to 90\% during the process of mask training. The real-valued mask is initialized using the soft-variant (Eq. \ref{eq:omp_init_soft}). This is because we find that the hard-variant is difficult to optimize with sparsity increase.

%%%%%%%%%%%%%%%%%%%%%%%%%%%%%%%%%%%%%%%%%%%%%%
\begin{figure}
    \parbox{0.25\linewidth}{
        \centering
        \includegraphics[width=1.0\linewidth]{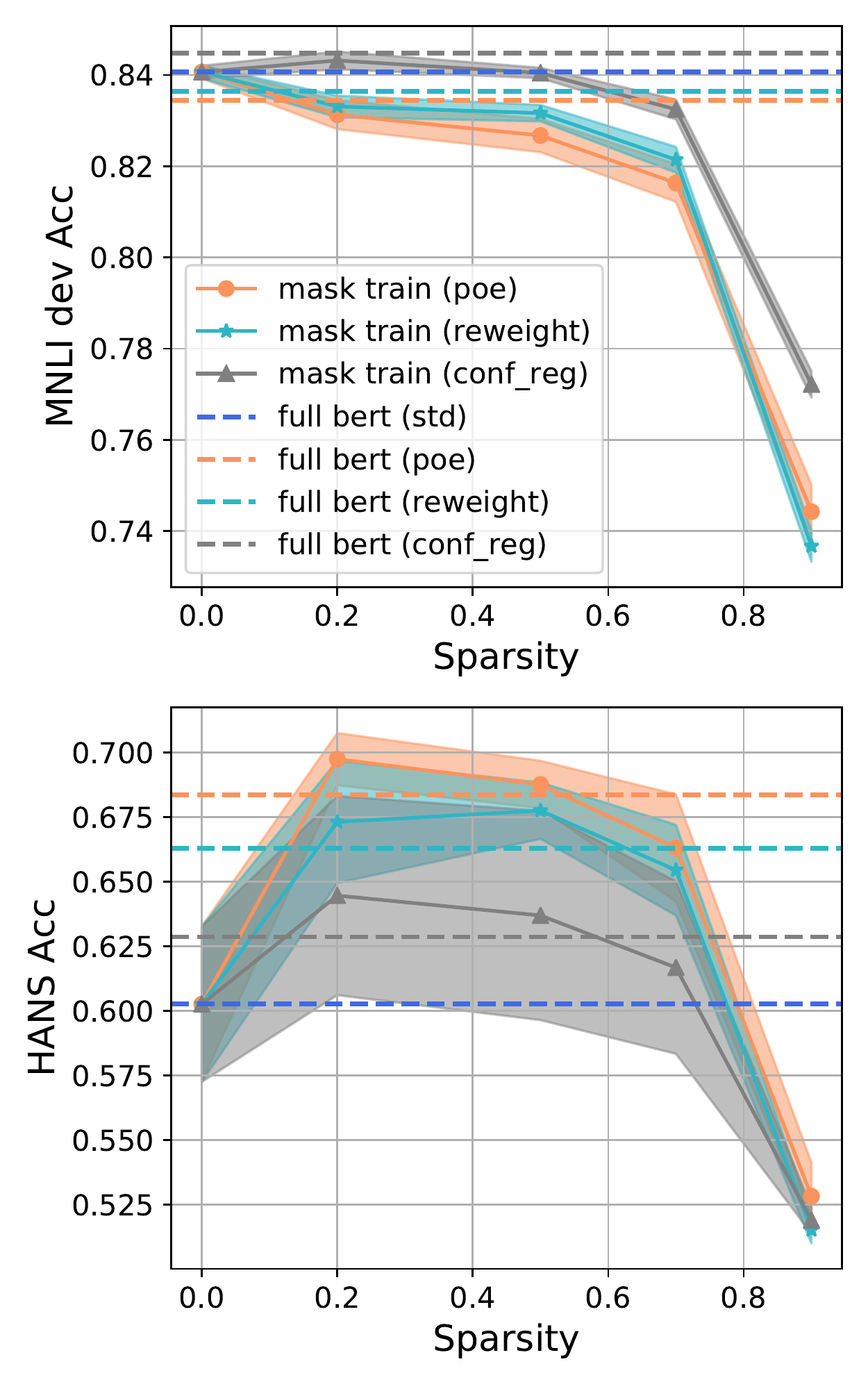}
    }
    \hfill
    \parbox{0.73\linewidth}{
        \centering
        \includegraphics[width=1.0\linewidth]{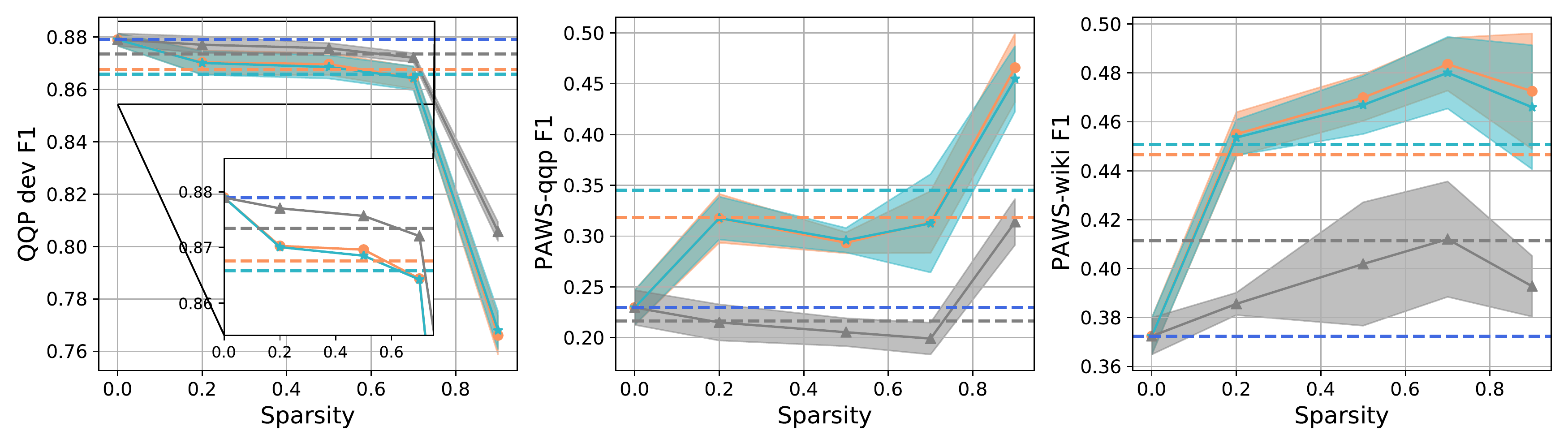}
        \includegraphics[width=1.0\linewidth]{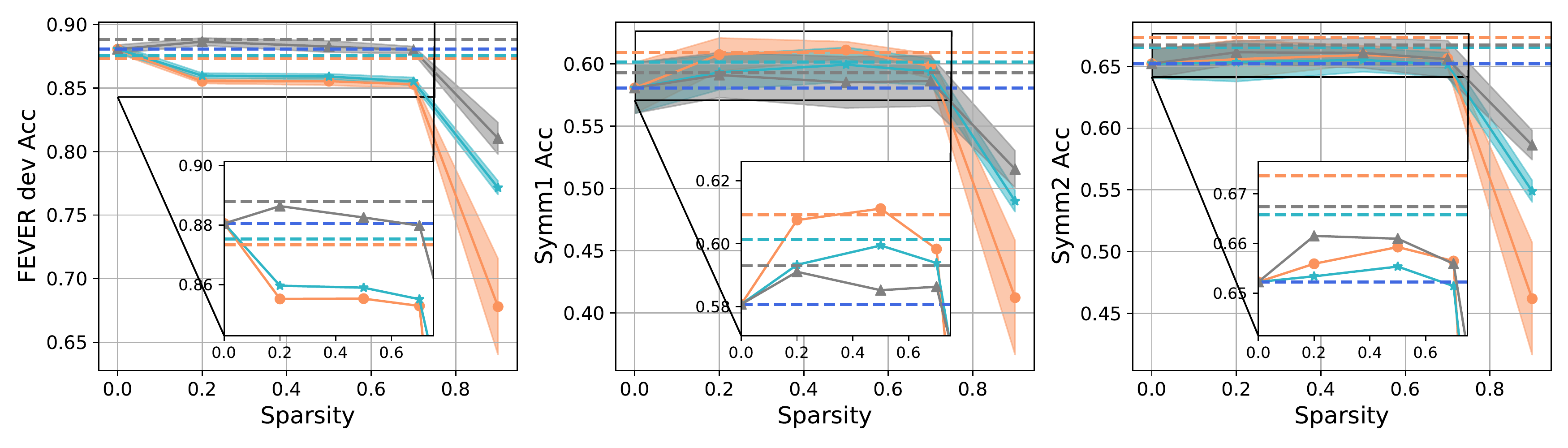}
    }
    \caption{Results of subnetworks pruned from the CE fine-tuned BERT, with different debiasing methods in pruning.}
    \label{fig:more_debias}
\end{figure}
%%%%%%%%%%%%%%%%%%%%%%%%%%%%%%%%%%%%%%%%%%%%%%
%%%%%%%%%%%%%%%%%%%%%%%%%%%%%%%%%%%%%%%%%%%%%%
\begin{figure}
    \parbox{0.25\linewidth}{
        \centering
        \includegraphics[width=1.0\linewidth]{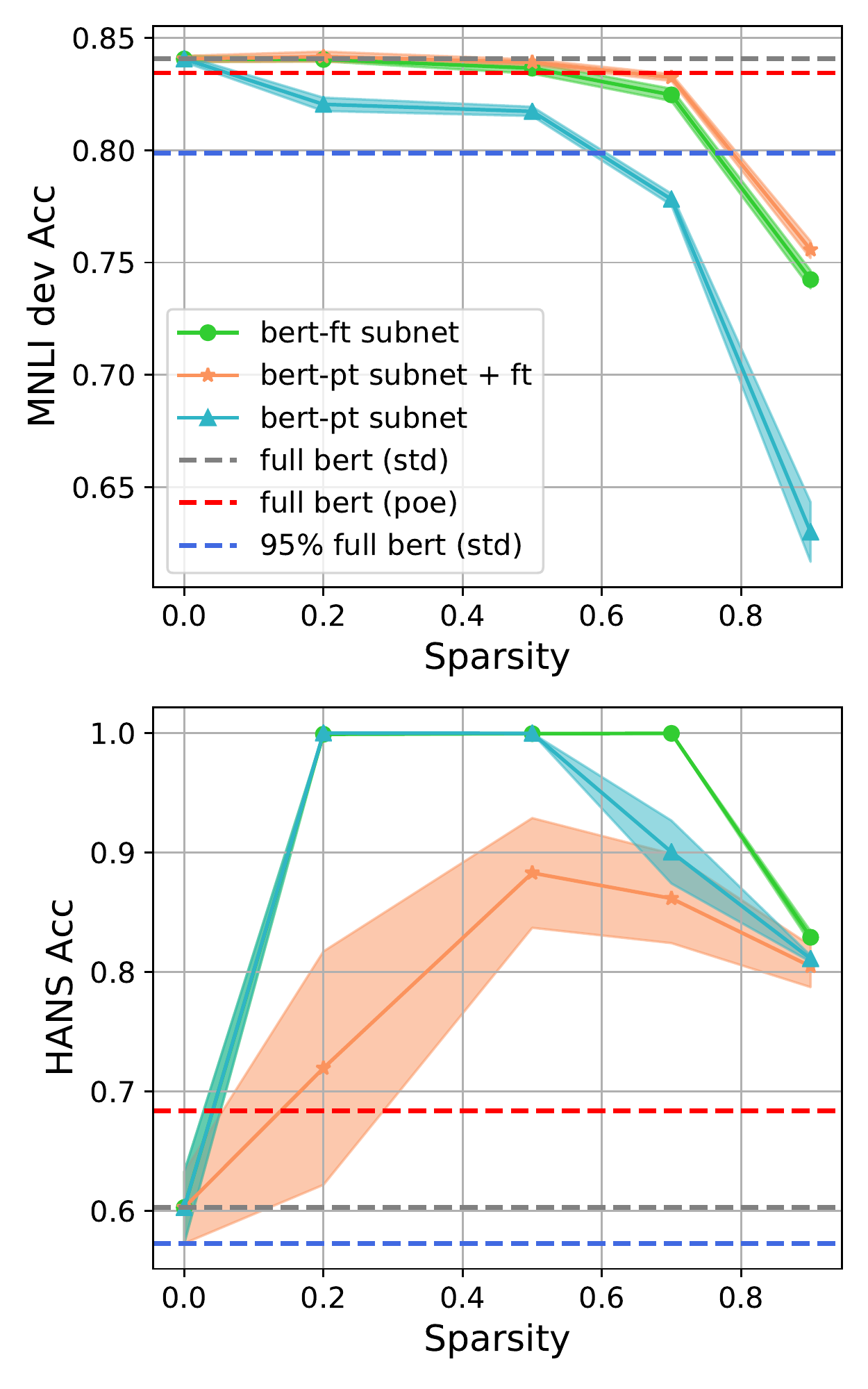}
    }
    \hfill
    \parbox{0.73\linewidth}{
        \centering
        \includegraphics[width=1.0\linewidth]{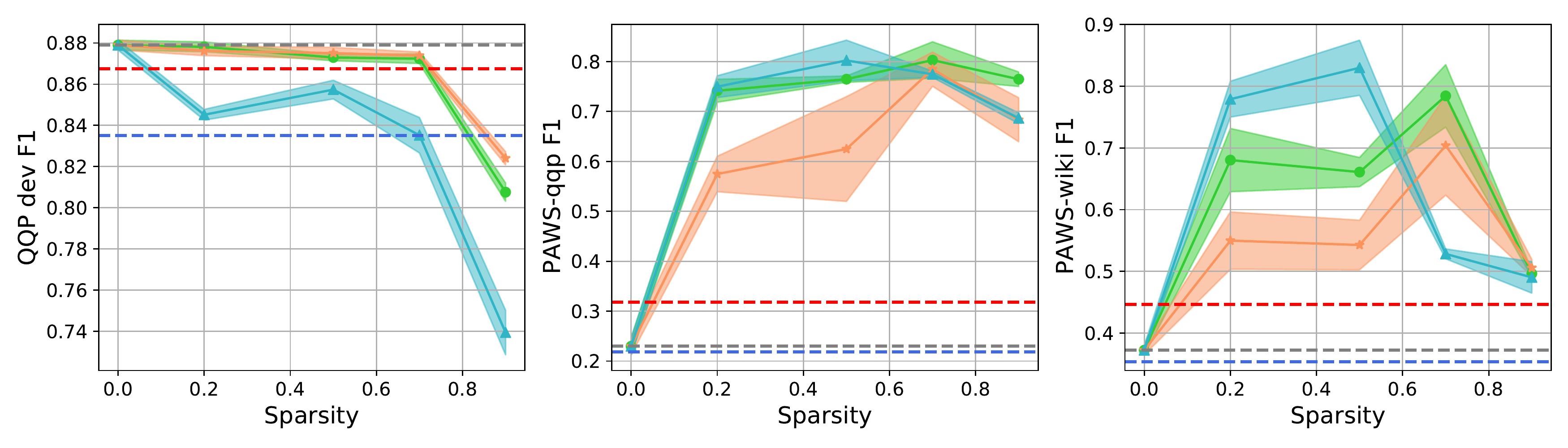}
        \includegraphics[width=1.0\linewidth]{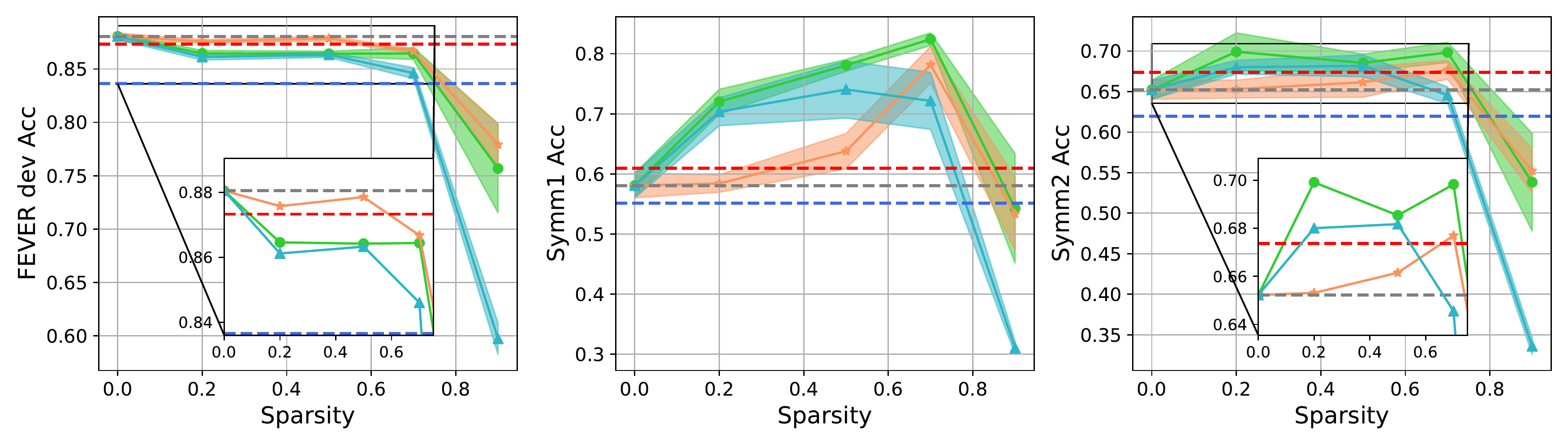}
    }
    \caption{Results of subnetworks found using the OOD information.}
    \label{fig:more_oracle}
\end{figure}
%%%%%%%%%%%%%%%%%%%%%%%%%%%%%%%%%%%%%%%%%%%%%%

\section{More Results and Analysis}
\subsection{More Debiasing Methods}
\label{appendix-c1}
In Section \ref{sec:srnets}, we mainly experiment with the PoE debiasing method. Here, we combine mask training with the other two debiasing methods, namely example reweighting and confidence regularization, and search for SRNets from the CE fine-tuned BERT. Fig. \ref{fig:more_debias} presents the results. As we can see: (1) Pruning with different debiasing methods almost consistently improves the OOD performance over the CE fine-tuned BERT. (2) The confidence regularization method (the grey lines) only achieves mild OOD improvement over the full BERT, while it preserves more ID performance compared with the other two methods. This phenomenon is in accordance with the results from \cite{MindTradeOff}, which propose the confidence regularization method to achieve a better trade-off between the ID and OOD performance.

\subsection{Sparse and Unbiased Subnetworks}
\label{appendix-c2}
Fig. \ref{fig:more_oracle} shows the results of mask training with the OOD training data. We can see that the general patterns in paraphrase identification and fact verification datasets are basically the same as the NLI datasets. Although the identified subnetworks cannot achieve 100\% accuracy on  PAWS and FEVER-Symmetric as on HANS, they substantially narrow the gap between OOD and ID performance, as compared with the full BERT. An exception is on the Symm2, where the upper bound of SRNets seems not very high. This is probably because we do not have enough examples (708 in total) to represent the data distribution of the FEVER-Symmetric dataset. Therefore, we conjecture that the existence of sparse and unbiased subnetworks might be ubiquitous.

\subsection{The Timing to Start Searching SRNets}
\label{appendix-c3}
\begin{figure}
    \parbox{0.26\linewidth}{
        \centering
        \includegraphics[width=1.0\linewidth]{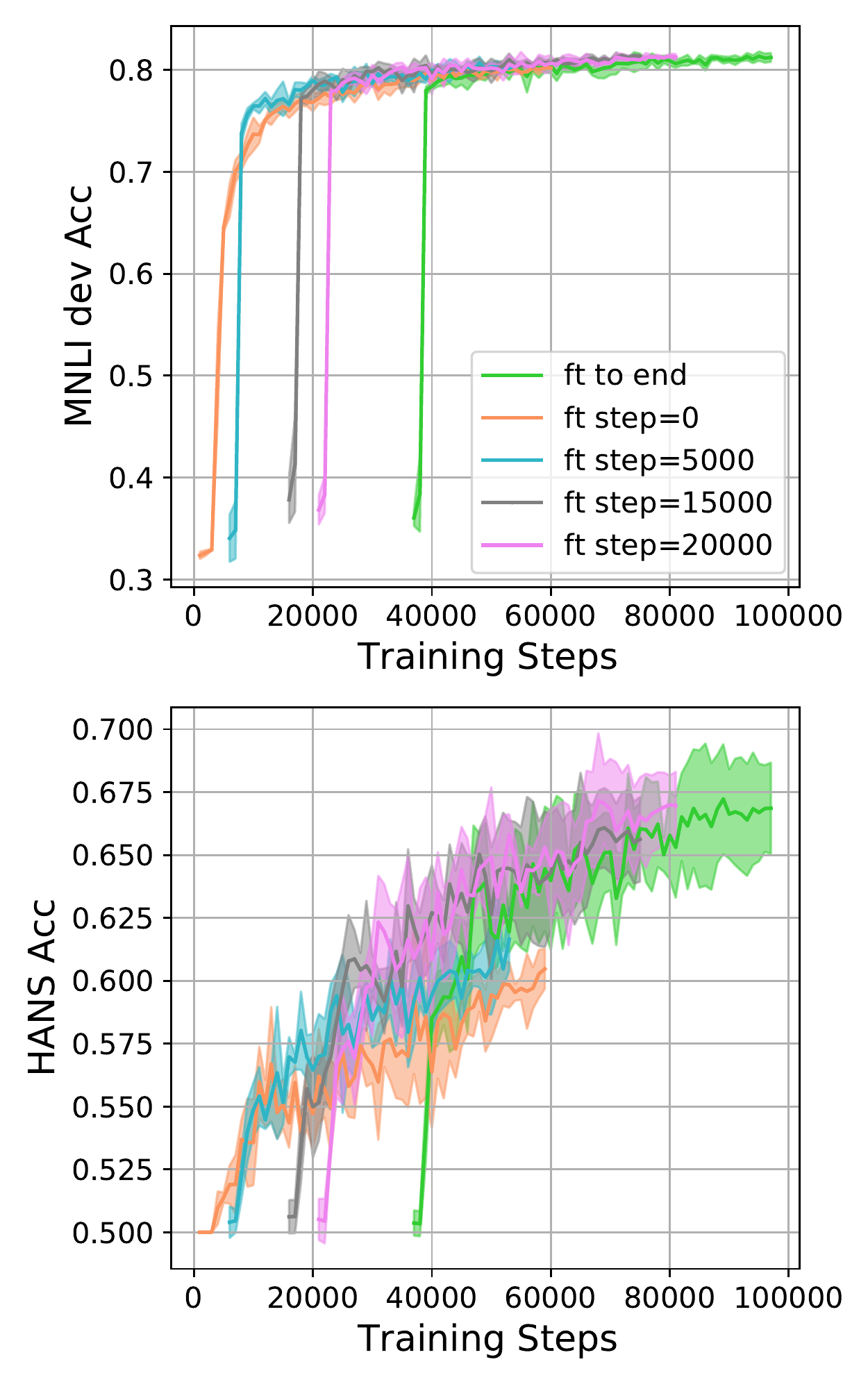}
    }
    \hfill
    \parbox{0.73\linewidth}{
        \centering
        \includegraphics[width=1.0\linewidth]{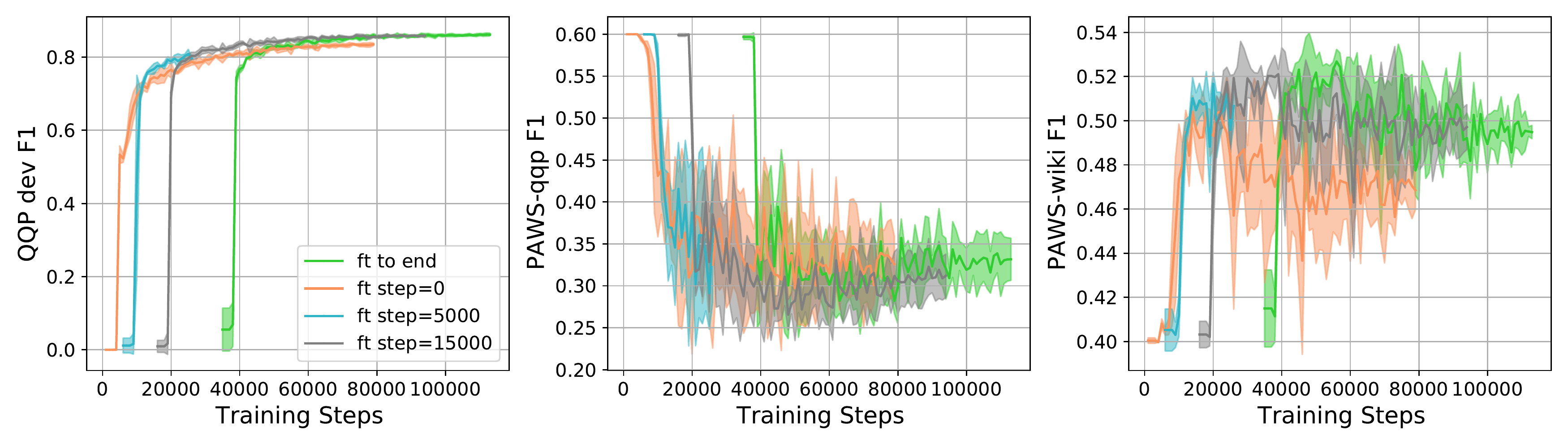}
        \includegraphics[width=1.0\linewidth]{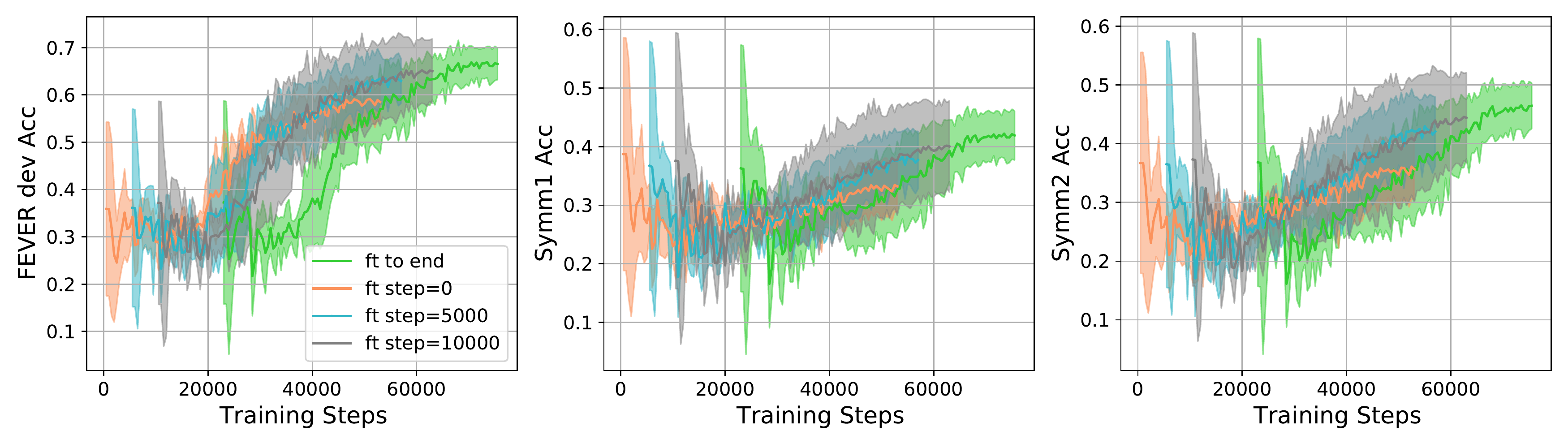}
    }
    \caption{Mask training curves starting from full BERT checkpoints fine-tuned for varied steps. The sparsity levels are 70\%, 70\% and 90\% for MNLI, QQP and FEVER respectively. At these sparsity levels, the gap between ``ft step=0'' and ``ft to end'' is the largest, according to Fig. \ref{fig:no_ft}.}
    \label{fig:prune_timing}
\end{figure}

Fig. \ref{fig:prune_timing} shows the mask training curves on all the 8 datasets. Similar to the NLI datasets, mask training on the other two tasks can achieve comparable results as ``ft to end'' by starting from an intermediate checkpoint of BERT fine-tuning. For QQP, we can start from 15,000 steps of full BERT fine-tuning (44\% of $t_{max}$). For FEVER, we can start from 10,000 steps (44\% of $t_{max}$).

%%%%%%%%%%%%%%%%%%%%%%%%%%%%%%%%%%%%%%%%%%%%%%
%%%%%%%%%%%%%%%%%%%%%%%%%%%%%%%%%%%%%%%%%%%%%%

\subsection{Ablation Studies on Gradual Sparsity Increase}
\label{appendix-c4}
As we mentioned in Appendix \ref{appendix-b3.3}, we increase the sparsity from 70\% to 90\% and adopt the soft variant of mask initialization. To explain the reason for using this specific strategy, we present the ablation study results in Tab. \ref{tab:gradual_sparse_ablation}. We can observe that: (1) Replacing the hard variant of mask initialization with the soft variant is beneficial, which leads to obvious improvements on the QQP, FEVER, Symm1 and Symm2 datasets. (2) Gradually increasing the sparsity further promotes the performance, with the 0.7$\sim$0.9 strategy achieving the best results on 7 out of the 8 datasets.

%%%%%%%%%%%%%%%%%%%%%%%%%%%%%%%%%%%%%%%%%%%%%%
\begin{table}[t]
    \caption{Ablation studies of the gradual sparsity increase schedule. The number of training epochs are 3, 5 and 5 for MNLI, QQP and FEVER respectively. The subnetworks are at $90\%$ sparsity. The numbers in the subscripts are standard deviations.}
    \label{tab:gradual_sparse_ablation}
    \parbox{0.25\linewidth}{
        \centering
        \resizebox{1.0\hsize}{!}{$
            \begin{tabular}{@{}l l c c c @{}}
                \toprule
                &                                  &MNLI   &HANS     \\\midrule
                \multirow{2}{*}{fixed}
                                     &hard         &72.09$_{0.92}$     &52.56$_{0.92}$      \\
                                     &soft       &72.63$_{0.31}$     &52.82$_{0.47}$        \\ \midrule
                \multirow{3}{*}{gradual}
                                     &0.2$\sim$0.9         &73.61$_{0.28}$   &53.90$_{0.87}$      \\
                                     &0.5$\sim$0.9       &75.06$_{0.31}$      &54.99$_{1.28}$        \\
                                     &0.7$\sim$0.9         &\textbf{76.84}$_{0.46}$      &\textbf{56.72}$_{0.75}$     \\
                \bottomrule
                \end{tabular}
        $}
    }
    \hfill
    \parbox{0.36\linewidth}{
        \centering
        \resizebox{1.0\hsize}{!}{$
            \begin{tabular}{@{}l l c c c c @{}}
                \toprule
                &           &QQP   &$\text{PAWS}_{qqp}$ &$\text{PAWS}_{qqp}$     \\\midrule
                \multirow{2}{*}{fixed}
                                     &hard         &71.64$_{1.85}$   &\textbf{55.70}$_{1.92}$   &49.59$_{1.84}$  \\
                                     &soft       &77.08$_{0.66}$   &46.48$_{3.55}$   &49.38$_{0.98}$  \\ \midrule
                \multirow{3}{*}{gradual}
                                     &0.2$\sim$0.9         &75.79$_{0.39}$   &51.57$_{0.69}$   &47.94$_{0.98}$  \\
                                     &0.5$\sim$0.9       &77.54$_{0.47}$   &50.92$_{0.97}$   &48.86$_{0.89}$  \\
                                     &0.7$\sim$0.9       &\textbf{79.49}$_{0.58}$   &46.59$_{1.81}$   &\textbf{51.15}$_{0.73}$  \\
                \bottomrule
                \end{tabular}
        $}
    }
    \hfill
    \parbox{0.36\linewidth}{
        \centering
        \resizebox{1.0\hsize}{!}{$
            \begin{tabular}{@{}l l c c c c @{}}
                \toprule
                &                       &FEVER   &Symm1    &Symm2     \\\midrule
                \multirow{2}{*}{fixed}
                                     &hard         &49.56$_{5.09}$   &27.45$_{2.94}$ &29.75$_{4.40}$ \\
                                     &soft       &72.80$_{0.95}$    &46.67$_{0.73}$ &52.33$_{0.75}$ \\ \midrule
                \multirow{3}{*}{gradual}
                                     &0.2$\sim$0.9         &73.53$_{1.36}$   &46.47$_{1.66}$ &52.42$_{1.39}$   \\
                                     &0.5$\sim$0.9       &77.01$_{0.43}$   &49.87$_{0.95}$ &56.57$_{0.22}$   \\
                                     &0.7$\sim$0.9       &\textbf{79.01}$_{0.68}$   &\textbf{51.74}$_{0.71}$   &\textbf{58.17}$_{0.33}$  \\
                \bottomrule
                \end{tabular}
        $}
    }
\end{table}
%%%%%%%%%%%%%%%%%%%%%%%%%%%%%%%%%%%%%%%%%%%%%%

\subsection{Results on RoBERTa-base and BERT-large}
\label{appendix-c5}
It has been shown by \cite{HendrycksLWDKS20,EmpiricalStudyRobustness} that pre-trained model RoBERTa \cite{RoBERTa} have better OOD generalization than BERT. \cite{EmpiricalStudyRobustness} also shows that larger PLMs, which are more computationally expensive, are more robust. To examine whether our conclusions can generalize to RoBERTa and larger versions of BERT, we conduct mask training on the standard fine-tuned RoBERTa-base and BERT-large models and use the PoE debiasing loss in the mask training process.

The results are shown in Tab. \ref{tab:bert-large-roberta}. We can see that, for RoBERTa-base: (1) At 50\% sparsity, the searched subnetworks outperform the full RoBERTa (std) by 6.84 points on HANS, with a relative small drop of 1.74 on MNLI, validating that SRNets can be found in RoBERTa. (2) At 70\% sparsity, the vanilla mask training produces subnetworks with undesirable ID performance and OOD performance comparable to full model (std). In comparison, when we gradually increase the sparsity level from 50\% to 70\%, the ID and OOD performance are improved simultaneously, demonstrating that gradual sparsity increase is also effective for RoBERTa.

When it comes to BERT-large, the conclusions are basically the same as BERT-base and RoBERTa-base: (1) We can find 50\% sparse SRNets from BERT-large using the original mask training. (2) Gradual sparsity increase is also effective for BERT-large. Additionally, we find that the original mask training exhibits high variance at 70\% sparsity because the training fails for some random seeds. In comparison, with gradual sparsity increase, the searched subnetworks have better performance and low variance.

%%%%%%%%%%%%%%%%%%%%%%%%%%%%%%%%%%%%%%%%%%%%%%
\begin{table}[t]
    \caption{Results of RoBERTa-base and BERT-large on the NLI task. We conduct mask training with PoE loss on the standard fine-tuned PLMs. ``0.5$\sim$0.7" denotes gradual sparsity increase. The numbers in the subscripts are standard deviations.}
    \label{tab:bert-large-roberta}
    \parbox{0.49\linewidth}{
        \centering
        \resizebox{1.0\hsize}{!}{$
            \begin{tabular}{@{}l l c c c @{}}
                \toprule
                RoBERTa-base&                     &MNLI   &HANS     \\\midrule
                \multirow{2}{*}{full model}
                                     &std                  &87.14$_{0.21}$     &68.33$_{0.88}$      \\
                                     &poe                  &86.56$_{0.18}$     &76.15$_{1.35}$        \\ \midrule
                \multirow{3}{*}{mask train}
                                     &0.5                  &85.40$_{0.14}$     &75.17$_{0.55}$      \\
                                     &0.7                  &83.48$_{0.29}$     &68.63$_{1.33}$        \\
                                     &0.5$\sim$0.7         &84.41$_{0.15}$     &71.95$_{1.23}$     \\
                \bottomrule
                \end{tabular}
        $}
    }
    \hfill
    \parbox{0.49\linewidth}{
        \centering
        \resizebox{1.0\hsize}{!}{$
            \begin{tabular}{@{}l l c c c @{}}
                \toprule
                BERT-large&                     &MNLI   &HANS     \\\midrule
                \multirow{2}{*}{full model}
                                     &std                  &86.84$_{0.13}$     &69.44$_{2.39}$      \\
                                     &poe                  &86.25$_{0.17}$     &76.27$_{1.55}$        \\ \midrule
                \multirow{3}{*}{mask train}
                                     &0.5                  &85.47$_{0.28}$     &75.40$_{0.64}$      \\
                                     &0.7                  &77.54$_{6.10}$     &60.19$_{7.56}$        \\
                                     &0.5$\sim$0.7         &84.83$_{0.26}$     &70.18$_{2.24}$     \\
                \bottomrule
                \end{tabular}
        $}
    }
\end{table}
%%%%%%%%%%%%%%%%%%%%%%%%%%%%%%%%%%%%%%%%%%%%%%

\section{Related Work on Model Compression and Robustness}
\label{appendix-d}
Some prior attempts have also been made to obtain compact and robust deep neural networks. We discuss the relationship and difference between these works and our paper from three perspectives:

\paragraph{Robustness Types}
There are various types of model robustness, including generalization to in-distribution unseen examples, robustness towards dataset bias \cite{BeeryHP18,HANS,PAWS,FeverSym} and adversarial attacks \cite{AdversarialAttack}, etc. Among the researches on model compression and robustness, adversarial robustness \cite{ADMM1,ADMM2,HYDRA,RobustScratchTickets,LoyaltyRobustnessOfBERT} and dataset bias robustness \cite{MRM,WhatDoCompressLMForget} are the most widely studied. In this paper, we focus on the dataset bias problem, which is more common than the worst-case adversarial attack, in terms of real-world application.

\paragraph{Compression Methods} A major direction in robust model compression is about the design of compression methods. \cite{TowardsCompactRobustDnn} investigate the effect of magnitude-based pruning on adversarially trained models. \cite{ADMM1,ADMM2} treat sparsity and adversarial robustness as a constrained optimization problem, and solve it using the alternating direction method of multipliers (ADMM) framework \cite{ADMM}. \cite{HYDRA,MRM,VulnerabilitySuppression} combine learnable weight mask (i.e., mask training) and robust training objectives. Our study investigates the use of magnitude-based pruning and mask training, which are also widely employed in the literature of BERT compression.

\paragraph{Application Fields}
Despite the topic of model compression and robustness has been proposed for years, it is mostly studied in the context of computer vision (CV) tasks and models, and few attention has been paid to the NLP field. Considering the real-world application potential of PLMs, it is critical to study the questions of PLM compression and robustness jointly. To this end, some recent studies extend the evaluation of compressed PLMs to consider adversarial robustness \cite{LoyaltyRobustnessOfBERT} and dataset bias robustness \cite{WhatDoCompressLMForget}. 

Although our work shares the same topic with \cite{WhatDoCompressLMForget}, we differ in several aspects. First, the scope and focus of our research questions are different. They aim at analyzing the impact of different compression methods (pruning and knowledge distillation \cite{KD}) on the OOD robustness of standard fine-tuned BERT. By contrast, we focus on subnetworks obtained from different pruning and fine-tuning paradigms and consider both standard fine-tuning and debiasing fine-tuning. Second, our conclusions are different. The results of \cite{WhatDoCompressLMForget} suggest that pruning generally has a negative impact on the robustness of BERT. In comparison, we revel the consistent existence of sparse BERT subnetworks that are more robust to dataset bias than the full model.

\section{More Discussions}
\subsection{How to Predict the Timing to Start Searching SRNets?}
\label{appendix-e1}
A feasible way of solution is to stop full BERT fine-tuning when there is no significant improvement across several consecutive evaluation steps. The patience of early-stopping can be determined based on the computational budget. If our resource is limited, we can at least directly training the mask on $\boldsymbol{\theta}_{pt}$, which can still produce SRNets at 50\% sparsity (as shown by Section \ref{sec:subnet_wo_ft_results}).

\subsection{How to Generalize to Other Scenarios?}
\label{appendix-e2}
In this work, we focus on NLU tasks and PLMs from the BERT family. However, the methodology we utilize is agnostic to the type of bias, task and backbone model. Theoretically, it can be flexibly adapted to other scenarios by simply change the spurious features to train the bias model (for the three debiasing methods considered in this paper) or combine the pruning method with another kind of debiasing method that also involves model training. In the future work, we would like to extend our exploration to other types of PLMs (e.g., language generation models like GPT \cite{GPT} and T5 \cite{T5}) and other types of NLP tasks (e.g., dialogue generation).

% Optionally include extra information (complete proofs, additional experiments and plots) in the appendix.
% This section will often be part of the supplemental material.

\end{document}